\pdfoutput=1
\documentclass[11pt]{article}

\usepackage[preprint]{acl}

\usepackage{times}
\usepackage{latexsym}

\usepackage[T1]{fontenc}
\usepackage[utf8]{inputenc}

\usepackage{microtype}

\usepackage{inconsolata}

\usepackage{graphicx}

\usepackage{booktabs}
\usepackage{enumitem}
\usepackage{algorithm}
\usepackage{algpseudocode}
\usepackage{amsmath} 
\usepackage{multirow}
\usepackage{subcaption}
\usepackage{lipsum} 
\usepackage{float}
\usepackage{siunitx}
\usepackage{colortbl}
\usepackage{graphicx}
\usepackage{subcaption}
\usepackage{tcolorbox}
\usepackage{setspace}

\definecolor{myblue}{HTML}{2069d6}
\definecolor{myred}{HTML}{db8840}
\definecolor{mypurple}{HTML}{893aba}
\newcommand{\vivid}[1]{\textcolor{myblue}{\textbf{\textit{#1}}}} 
\newcommand{\technical}[1]{\textcolor{myred}{\textbf{\textit{#1}}}} 
\newcommand{\note}[1]{\textcolor{mypurple}{\textbf{\textit{#1}}}} 

\definecolor{gongyaoColor}{RGB}{34, 139, 34}

\title{JRE-L: \underline{J}ournalist, \underline{R}eader, and \underline{E}ditor LLMs in the \underline{L}oop \\ for Science Journalism for the General Audience}
\author{Gongyao Jiang$^1$,  Xinran Shi$^1$,  Qiong Luo$^{1,2,}$\thanks{Corresponding Author}  \\
  $^1$The Hong Kong University of  Science and Technology (Guangzhou)\\
  $^2$The Hong Kong University of Science and Technology \\
  \texttt{jianggongyao@gmail.com, luo@cse.ust.hk}\\}

\begin{document}
\maketitle
\begin{abstract}
Science journalism reports current scientific discoveries to non-specialists, aiming to enable public comprehension of the state of the art.
This task is challenging as the audience often lacks specific knowledge about the presented research.
We propose a JRE-L framework that integrates three LLMs mimicking the writing-reading-feedback-revision loop. 
In JRE-L, one LLM acts as the journalist, another LLM as the general public reader, and the third LLM as an editor.
The journalist's writing is iteratively refined by feedback from the reader and suggestions from the editor.
Our experiments demonstrate that by leveraging the collaboration of two 7B and one 1.8B open-source LLMs, we can generate articles that are more accessible than those generated by existing methods, including prompting single advanced models such as GPT-4 and other LLM-collaboration strategies.
Our code is publicly available at \url{github.com/Zzoay/JRE-L}.

\end{abstract}

\section{Introduction}
Science journalism creates news articles that cover a wide range of scientific research, enhancing the public's understanding of science \citep{gopfert2008strength, allan2011introduction, angler2017science}.
With rapid advances in various disciplines, science journalism struggles to keep pace with the exponential growth of knowledge. 
In response, automatic science journalism (ASJ) has been proposed to expedite the filtering, learning, and communication of scientific knowledge \citep{dangovski2021we}.

\begin{figure}[t]
    \centering
    \begin{subfigure}[b]{0.17\textwidth}
        \centering
        \includegraphics[width=\textwidth]{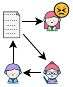} 
        \caption{}
        \label{fig:sub1}
    \end{subfigure}
    % \\
    \begin{subfigure}[b]{0.3\textwidth}
        \centering
        \includegraphics[width=\textwidth]{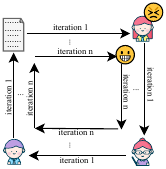} 
        \caption{}
        \label{fig:sub2}
    \end{subfigure}
    % \hfill
    % \begin{subfigure}[b]{0.19\textwidth}
    %     \centering
    %     \includegraphics[width=\textwidth]{fig/intro-c_cropped.pdf} 
    %     \caption{Inspired}
    %     \label{fig:sub3}
    % \end{subfigure}
    % \caption{An article written by a scientific journalist \raisebox{-0.8ex}{\includegraphics[width=0.025\textwidth]{fig/writer_icon.png}} is difficult for the general reader \raisebox{-0.8ex}{\includegraphics[width=0.025\textwidth]{fig/reader_icon.png}} to understand (a). Incorporating the reading feedback into a iterative journalism with the reviewing and suggesting by an editor \raisebox{-0.8ex}{\includegraphics[width=0.028\textwidth]{fig/editor_icon.png}} can help make the article easy for the general reader to understand.}
    \caption{An article written by a science journalist \raisebox{-0.8ex}{\includegraphics[width=0.025\textwidth]{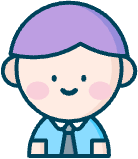}} may be challenging for the general reader \raisebox{-0.8ex}{\includegraphics[width=0.025\textwidth]{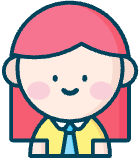}} without the reader's feedback to the editor \raisebox{-0.8ex}{\includegraphics[width=0.028\textwidth]{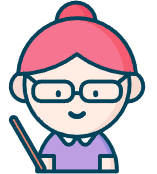}} in the revision cycle (a). Incorporating the reader's feedback into the journalism cycle can help enhance the readability of the article (b).}
    \label{fig:read}
\end{figure}

The essence of ASJ lies in elucidating complex technical content for readers, thereby facilitating their comprehension of advanced research \citep{cardenas2023don}. 
% However, ASJ-generated content can be challenging for the general audience who lack in-depth knowledge of specific fields. 
The degree to which content is embraced depends on the reader's domain knowledge \citep{august2024know}, thus scientific content may be challenging for the general reader, as illustrated in Figure \ref{fig:read} (a).
Some researchers have developed parallel corpora \citep{dangovski2021we, goldsack2022making, cardenas2023don}, where the target content is extracted from online scientific news or journals. 
However, these press releases often remain technical, likely because they are originally tailored for researchers rather than the general audience. 
% Models trained on such content struggle to generate materials easily understandable for a broader audience.
% \begin{figure}
%     \centering
%     \includegraphics[width=0.5\textwidth]{fig/complexity.pdf}
%     \caption{Reader experience varies with content technicality. Science journalism for the general audience demands high accessibility.}
%     \label{fig:com}
% \end{figure}
% \begin{figure}
%     \centering
%     \includegraphics[width=0.88\linewidth]{fig/readability.pdf}
%     \caption{Readability issues may pose challenges for scientific writers \raisebox{-0.8ex}{\includegraphics[width=0.03\textwidth]{fig/writer_icon.png}} to identify independently. Feedback from the general audience \raisebox{-0.8ex}{\includegraphics[width=0.03\textwidth]{fig/reader_icon.png}} can assist them in considering adjustments for higher readability.}
%     \label{fig:read}
% \end{figure}

Large language models (LLMs) have shown impressive proficiency in instruction adherence and content generation \citep{openai2024gpt4, bai2023qwen}, thereby making them potential tools for ASJ.
Furthermore, LLMs have exhibited social intelligence, enabling them to play realistic roles and collaborate in real-world tasks \citep{park2023generative, talebirad2023multi, qian2023communicative}.
Moreover, LLMs can iteratively improve their performance by context updates without training \citep{zhao2024expel, senel-etal-2024-generative}.
Motivated by these observations, we propose a communicative and iterative framework that leverages LLMs to gradually accomplish the ASJ task.

Our goal is to automatically generate a popular science article based on a technical paper and improve the accessibility of the generated article to the general audience. 
In real-world journalism, a journalist typically receives and learns from revision suggestions from a professional editor \citep{nip2006exploring, anderson2015post}, as reviews can lead to writing improvements \citep{bryant2002fluid, cho2011learning}.
Furthermore, in science communication, receiving feedback from the audience can improve scientists' communication skills \citep{brownell2013science, clark2016science}.
Thus, the introduction of general readers' feedback into the journalism loop is promising to make the writing more accessible to the general public, as illustrated in Figure \ref{fig:read} (b).

Based on real-world practices and research results, we design the JRE-L framework, in which three LLMs collaborate in a loop of writing, reading, feedback, and revision, to generate highly accessible popular science articles.
Concretely, we have an LLM serve as the journalist writing for readers who lack domain knowledge of the article. 
To help expose writing problems that might hinder the reading experience of general readers, we have another LLM, smaller than the journalist LLM, to serve as a general reader.
This reader LLM reads the article written by the journalist and takes notes for giving feedback.
As a less proficient model, the reader LLM needs material that is easily understandable to take comprehensive notes. 
Therefore, the more accessible the written article is, the greater the clarity and accuracy of the reader's notes will exhibit.

LLMs have shown the capability of evaluating the quality of text \citep{chan2023chateval, zheng2024judging, desmond2024evalullm}.
Therefore, we let an editor (the third LLM) evaluate the correctness and comprehensiveness of the reader's notes and then provide suggestions for the revision of the journalist's article. 
The journalist then revises the article based on the suggestions. 
By this iterative and parameter-free tuning process, the popular science article is enhanced and made more accessible to a general audience.
To the best of our knowledge, our work is the first study on LLM collaboration for ASJ.

To assess our proposed method, we employ both automatic metrics and human evaluation on measures including readability, information conveyance, authenticity, and interestingness of our generated articles. 
Compared with other methods, including those with fine-tuning and prompting on various LLMs, our proposed method achieves the highest readability while remaining competitive on the other measures. 
We also provide a detailed analysis, including ablation studies of removing the editor LLM, removing the reader LLM, or removing both, as well as trend analysis and case studies, to offer a comprehensive understanding on LLMs in the ASJ task.
% The code will be made publicly available for research purposes.

In brief, we make the following contributions: 
\begin{itemize}[noitemsep, topsep=0pt] 
    \item An ASJ framework with collaborative LLMs, generating content of high readability.
    \item Comprehensive experiments, analysis, and insights for applying LLMs in ASJ. 
\end{itemize}

\section{Related Work}

\noindent \textbf{Automatic Science Journalism.} 
In recent years, some researchers have explored the application of LLM on authoring scientific articles \citep{wang2024scimon, baek2024researchagent, wang2024auto}.
% Particularly, ASJ has gained increasing interest.
\citet{dangovski2021we} created a parallel corpus and provided a sequence-to-sequence method to generate news summaries from scientific articles.
% However, this dataset is not available to the public because of the licensing restriction.
\citet{goldsack2022making} released two corpora, focusing on the biomedical and life science domains.
% Similarly, they employed a standard sequence-to-sequence model for such tasks.
\citet{cardenas2023don} constructed a dataset in various scientific fields and integrated the discourse structure of papers with their metadata to guide the generation.
These methods of fine-tuning on small models can provide a good match with the reference, but there is still room for improvement in readability.
In this work, we present an approach that integrates LLMs as agents to iteratively enhance readability.

\begin{figure*}[t]
    \centering
    \includegraphics[width=1\textwidth]{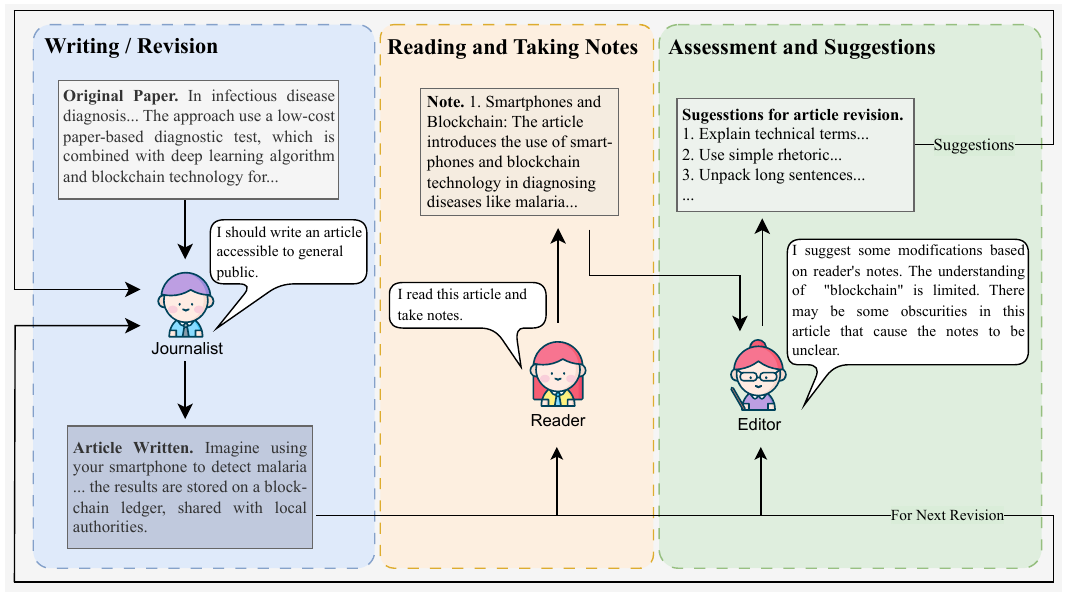}
    \caption{Overview of our iterative ASJ framework, JRE-L.}
    \label{fig:overview}
\end{figure*}

\noindent \textbf{Large Language Models.} 
% With the emergence of the LLM, advanced models \citep{openai2024gpt4, touvron2023llama, anthropic2023claude, jiang2023mistral, bai2023qwen} have shown performance comparable to that of humans in a variety of real-world tasks.
Our study involves three aspects related to LLMs.
First, various studies show strong abilities of LLMs in writing scientific content \citep{pu2024scinews, kumar2024longform, lee2024design}.
Second, LLMs have demonstrated remarkable intelligence in social simulation \citep{park2023generative, ziems2024can} and real-world tasks \citep{liu2023dynamic, chen2023gamegpt, ding2023designgpt, qian2023communicative, qin2024tool}.
% \citet{park2023generative} leveraged LLMs for social simulation, showing communication and collaboration between LLM-based agents.
% \citet{liu2023dynamic} built an LLM-collaboration architecture for enhancing reasoning and code generation tasks.
% \citet{qian2023communicative} used different LLM-based agents throughout the development process for software engineering.
Inspired by this collection of work, we utilize LLMs as communicative agents to make content accessible to the general audience through a process resembling real-world practice.
Third, previous studies have demonstrated that LLMs can be iteratively optimized by in-context learning without internal parameter tuning \citep{yang2023failures, zhao2024expel, senel-etal-2024-generative, chenboosting2024}.
Thus, this work proposes a parameter-free tuning framework that iteratively improves writing through interactions among LLMs.

\section{Methodology}
Following ASJ (\citet{dangovski2021we, goldsack2022making, cardenas2023don}) aims to automatically distill a scientific paper into an article accessible to a broader audience.
Our ASJ framework JRE-L employs an iterative workflow of writing, reading, suggestion-making, and revision among three LLMs, as illustrated in Figure \ref{fig:overview}.
All prompts for each LLM agent are listed in Appendix \ref{sec:prompts}.

\subsection{The LLM Journalist}
LLMs have shown strong writing abilities \citep{yuan2022wordcraft, wasi2024llms}. 
Thus, they are promising tools for rewriting a given paper into a more accessible version.
Following established strategies \citep{zheng2024judging, zhang2024llm}, we start with prompting an LLM to assume the role of a journalist.
Subsequently, the LLM journalist $\mathcal{J}$ is prompted that, given the paper $x$, its task is to compose an article $p_0$ for the general public.
\begin{equation}
    p_0 = \mathcal{J} \left( x \right)
\end{equation}
where journalist $\mathcal{J}$ is initialized from an LLM by a task prompt, one-shot demonstration, and fine-tuning.

\subsection{The LLM Reader}

\begin{figure}[t]
    \centering
    \includegraphics[width=0.482\textwidth]{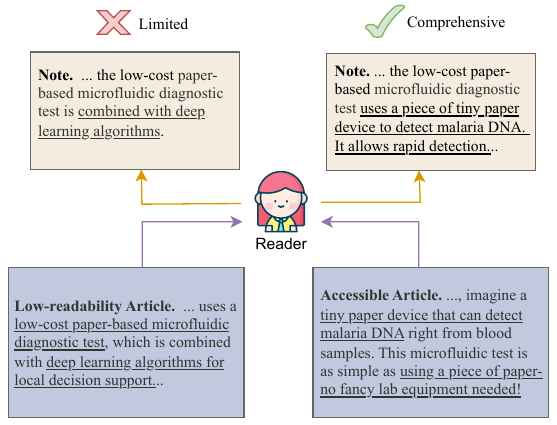}
    \caption{Accessible content helps the reader take comprehensive notes.}
    \label{fig:student}
\end{figure}

In our initial attempts, we asked an LLM to directly assess the readability of the generated article. 
However, the results were unsatisfactory, probably due to the gap between human and model perceptions of reading difficulty.
As illustrated in the two text boxes at the bottom of Figure \ref{fig:student}, LLMs regarded both pieces of writing at a similar level of readability, as they all incorporated essential information, even if the terminology ``low-cost paper-based microfluidic diagnostic tests'' on the left side was not clearly explained.
However, a human reader perceived the writing on the right is as more accessible.\footnote{We briefly document other unsuccessful attempts in Appendix \ref{sec:fail}.} 

To address this readability assessment problem, we design a separate reader LLM to read the content and generate notes. Our idea is inspired by the accumulation of errors, a common phenomenon in pipeline systems \citep{caselli2015s, wu2018beyond, dziri2024faith}. 
Specifically, we utilize the propagation from the textual readability of the journalist's article to the reader's comprehension in the writing-reading pipeline to induce the readability of the journalist's generated article.

Different from the LLM journalist, the reader LLM is of a smaller scale and thus has weaker reading comprehension skills, simulating a general reader with limited domain knowledge.
Once presented with the article crafted by the journalist, the reader LLM is employed to read the article and take notes.
Specifically, we instruct the reader LLM to explain key terms in the article by extracting the explanations, if present, directly from the article or offering explanations for these terms, otherwise.
\begin{equation}
\label{eq:note}
    r_i = \mathcal{R} \left(p_{i-1} \right)
\end{equation}
where $r_i$ is the notes taken by the reader $\mathcal{R}$ on the current version of writing from the journalist $\mathcal{J}$, $i = 1,2,\ldots,n$, and $n $ is the number of iterations.

Intuitively, if the article is more accessible, the reader's notes will be more comprehensive.
For instance, in Figure \ref{fig:student}, when the piece (left) lacks a detailed explanation of the term ``paper-based microfluidic diagnostic test'', the reader will only note this term as being ``combined with deep learning''. 
When the article (right) explains the usage and advantages of this text in more detailed and plain language, the reader LLM produces better notes.
Through this readability propagation from the journalist's article to the comprehensiveness of the reader's notes, the editor LLM can better recognize issues in the journalist's writing and then provide suitable suggestions for revision.

\subsection{Automated Suggestions and Revisions}
LLMs have demonstrated strong capabilities in serving as evaluators, widely utilized in various generative tasks \citep{chan2023chateval, zheng2024judging, desmond2024evalullm}.
Therefore, we employ an LLM as an editor for automated evaluation of reader comprehension and providing recommendations for article enhancement.
Given the article from the journalist and notes from the reader, the LLM editor $\mathcal{E}$ is tasked with assessing the quality of the reader's notes $r$ and identifying issues in the journalist's writing $p_{i-1}$ that may lead to reading obstacles. 
Next, the editor $\mathcal{E}$ offers suggestions $s_i$ for the journalist's content revision.
 \begin{equation}
    s_i = \mathcal{E} \left( p_{i-1}, r_i\right)
\end{equation}

For example, in Figure \ref{fig:overview}, the editor finds that the reader's understanding of the term ``blockchain'' is limited, possibly due to an insufficient explanation in the reading material. 
To address this perceived issue, the editor suggests that the article should ``explain technical terms.'' 
These suggestions are then incorporated into the instructions that will guide the journalist in revising the article.
Subsequently, with the strong ability to follow instructions, the journalist LLM $\mathcal{J}$ rewrites the article based on the suggestions:
\begin{equation}
    p_i = \mathcal{J} \left(x, p_{i-1}, s_i\right)
\end{equation}

Then, the revised piece is fed to the reader for reading and taking notes (Equation \ref{eq:note}) to continue the process. 
By an iterative cycle encompassing writing, note-taking, suggesting modifications, and revision among three LLMs, the article tailored for the general readership undergoes steady enhancement.
Algorithm \ref{alg:proc} presents this process.
% We analyze this process in Section \ref{sec:ana}.
\begin{algorithm}[t]
\small
\begin{algorithmic}[1]
\Statex \textbf{INPUT:} a scientific paper $\boldsymbol{x}$; journalist $\mathcal{J}$; reader $\mathcal{R}$; editor $\mathcal{E}$; number of iterations $n$.
\Statex \textbf{OUTPUT:} a news article $\boldsymbol{p}_n$ for the general audience.
\State $\boldsymbol{p}_0 \gets \mathcal{J}(\boldsymbol{x})$ \Comment{Initial writing}
\For {$i=1$ to $n$}
    \State $\boldsymbol{r}_i \gets \mathcal{R}\left(\boldsymbol{p}_{i-1}\right)$   \Comment{Reader's notes}
    \State $\boldsymbol{s}_i \gets \mathcal{E}\left(\boldsymbol{p_{i-1}}, \boldsymbol{r}_i\right)$  \Comment{Editor's suggestions}
    \State $\boldsymbol{p}_i \gets \mathcal{J}\left(\boldsymbol{x}, \boldsymbol{p_{i-1}}, \boldsymbol{s}_i\right)$ \Comment{Revision}
\EndFor
\State \textbf{Return} $\boldsymbol{p}_n$
\end{algorithmic}
\caption{\small JRE - L for ASJ}
\label{alg:proc}
\end{algorithm}

\section{Experiments}

\subsection{Setting}
\noindent \textbf{Datasets.}
We use three publicly available corpora in different disciplines as benchmarks, namely \emph{SCITech} \citep{cardenas2023don}, \emph{eLife}, and \emph{PLOS} \citep{goldsack2022making}.
For a fair comparison, the data split strategy is the same as that in these previous studies.
Appendix \ref{sec:hyper} briefly introduces these datasets.

\begin{table*}[t]
\centering
\small
\setlength{\tabcolsep}{4pt}
\begin{tabular}{lccccccccccc}
\toprule
                                               & \multicolumn{3}{c}{\textbf{SCITech}}                                                                                 & \multicolumn{3}{c}{\textbf{eLife}}                                                                                  & \multicolumn{3}{c}{\textbf{PLOS}}                                                                                    &                                       \\ \cmidrule(lr){2-4} \cmidrule(lr){5-7} \cmidrule(lr){8-10}
\multirow{-2}{*}{\textbf{Approach}}            & \textbf{CLI}$\downarrow$                        & \textbf{FKGL}$\downarrow$                      & \textbf{DCRS}$\downarrow$                       & \textbf{CLI}$\downarrow$                    & \textbf{FKGL}$\downarrow$                   & \textbf{DCRS}$\downarrow$                    & \textbf{CLI}$\downarrow$                    & \textbf{FKGL}$\downarrow$                    & \textbf{DCRS}$\downarrow$                 & \multirow{-2}{*}{\textbf{Avg.}$\downarrow$ }     & \multirow{-2}{*}{\textbf{Impr.(\%)}$\uparrow$ }          \\ 
\midrule
{\color[HTML]{485aa3} LLaMA-2-7B}     & 15.13                                 & 13.79                                 & 10.38                                & 15.16                                 & 14.03                                & 10.50                                & 15.36                                 & 14.28                                 & 10.54                                & 13.24             &     19.26$^{\dagger\dagger}$               \\
% {\color[HTML]{485aa3} Gemma-7B}       & 14.93                                 & 13.75                                 & 10.52                                & 15.01                                 & 12.08                                & 11.03                                & 15.52                                 & 12.29                                 & 10.92                                & 12.89                                 \\
{\color[HTML]{485aa3} Mistral-7B}    & 14.90                                 & 13.54                                 & 10.82                                & 14.61                                 & 11.72                                & 10.85                                & 15.38                                 & 11.98                                 & 11.21                                & 12.78              &   16.35$^{\dagger\dagger}$                \\
{\color[HTML]{485aa3} Qwen-1.5-7B}    & 14.77                                 & 13.50                                 & 10.72                                & 14.72                                 & 11.83                                & 10.92                                & 15.06                                 & 11.94                                 & 11.09                                & 12.73              &      16.03$^{\dagger\dagger}$             \\
{\color[HTML]{485aa3} $\text{Qwen-1.5-7B}_{OS}$}    & 14.44 & 13.51 & 10.31 & 14.22 & 11.62 & 10.52 & 14.86 & 11.43 & 10.82 & 12.41   &  13.86$^{\dagger\dagger}$             \\
{\color[HTML]{485aa3} LLaMA-3-8B}     & 14.84                                 & 13.18                                 & 10.41                                & 14.55                                 & 11.65                                & 10.49                                & 15.18                                 & 12.01                                 & 10.88                                & 12.58            &   15.02$^{\dagger\dagger}$                    \\
{\color[HTML]{485aa3} $\text{LLaMA-3-8B}_{OS}$}     & 14.65 & 13.01 & 10.21 & 14.13 & 11.57 & 10.35 & 14.76 & 11.81 & 10.79 & 12.36    &   13.51$^{\dagger\dagger}$                \\
{\color[HTML]{485aa3} Mixtral-8x7B}   & 13.98                                 & 13.25                                 & 10.36                                & 14.21                                 & 12.01                                & 10.28                                & 15.34                                 & 11.58                                 & 10.98                                & 12.44               &  14.07$^{\dagger\dagger}$                 \\
{\color[HTML]{485aa3} Qwen-1.5-72B}   & 13.78                                 & 13.10                                 & 10.25                                & 14.17                                 & 12.09                                & 10.35                                & 15.18                                 & 11.75                                 & 10.62                                & 12.37              &    13.58$^{\dagger\dagger}$                 \\
{\color[HTML]{485aa3} GPT-3.5-Turbo}  & 14.98                                 & 13.62                                 & 10.81                                & 14.35                                 & 11.87                                & 10.98                                & 15.11                                 & 11.92                                 & 10.87                                & 12.72               &    15.96$^{\dagger\dagger}$                \\
{\color[HTML]{485aa3} GPT-4}          & 13.48                                 & 12.13                                 & 10.14                                & 13.96                                 & 10.87                                & 10.11                                & 14.86                                 & 11.78                                 & 10.47                                & 11.98             &   10.77$^{\dagger\dagger}$                   \\
{\color[HTML]{d17630} $\text{BART}_{FT}$}           & 13.43                                 & 15.22                                 & 10.66                                & 12.32                                 & 10.65                                & 9.19                                 & 15.61                                 & 14.24                                 & 10.51                                & 12.43      &  14.00$^{\dagger\dagger}$                           \\
{\color[HTML]{d17630} $\text{Qwen}_{FT}$} & 13.37                                 & 14.79                                 & 10.48                                & 12.15                                 & 10.63                                & \underline{9.12}                                 & 15.54                                 & 13.95                                 & 10.58                                & 12.29      &       13.02$^{\dagger\dagger}$                      \\
{\color[HTML]{d17630} $\text{LLaMA}_{FT}$} & 13.31 & 14.52 & 10.12 & 12.01 & 10.31 & 9.14 & 15.03 & 13.64 & 10.37 & 12.05  &     11.29$^{\dagger\dagger}$              \\
{\color[HTML]{48968B} CollabStory}    & 13.82 & 12.13 & 10.32 & 13.41 & 11.43 & 9.73 & 14.30 & 11.70 & 10.11 & 11.88     &   10.02$^{\dagger\dagger}$                            \\
{\color[HTML]{48968B} ChatDev}    & 13.51 & 12.21 & 10.49 & 13.07 & 10.92 & 9.77 & 13.91 & 10.97 & 9.92 & 11.64 & \ \ 8.16$^{\dagger}$                            \\
{\color[HTML]{754494} \textbf{$\text{JRE-L}_{OS}$}}  &   \underline{12.74} & \underline{10.37} & \underline{9.89} & \underline{11.86} & \underline{10.08} & 9.26 & \underline{12.84} & \underline{10.03} & \underline{9.87} & \underline{10.77} &  0.74 \\
{\color[HTML]{754494} \textbf{$\text{JRE-L}_{FT}$}}   & 12.94                                 & 13.33                                 & 10.33                                & 12.04                                 & {\textbf{9.85}} & {\textbf{9.04}} & 13.15                                 & 11.48                                 & 10.17                                & 11.37       &     \ \    5.98$^{\dagger}$                   \\
\rowcolor{gray!15} {\color[HTML]{754494} \textbf{JRE-L}}      & {\textbf{12.69}} & {\textbf{10.16}} & {\textbf{9.79}} & {\textbf{11.60}} & 10.10                                & 9.46                                 & {\textbf{12.74}} & {\textbf{10.00}} & {\textbf{9.69}} & {\textbf{10.69}} & 0.00 \\
% \midrule
% Reader $\rightarrow$ 7B  & 12.81 & \underline{10.35} & \textbf{9.68} & \underline{11.82} & \underline{10.01} & 9.51 & \textbf{12.67} & \textbf{9.93} & \underline{9.78} & \underline{10.73} \\
% $-$ Reading	& 13.21 & 10.63 & 10.33 & 12.22 & 10.78 & 10.02 & 13.35 & 10.59 & 10.25 & 11.26 \\
% $-$ Suggestions	& 13.25 & 10.69 & 10.39 & 12.17 & 10.83 & 10.08 & 13.31 & 10.74 & 10.42 & 11.32 \\
% $-$ Collaboration	& 13.50 & 11.01 & 10.71 & 12.47 & 10.99 & 10.41 & 13.65 & 10.91 & 10.70 & 11.59 \\
\midrule
Paper Abstracts                            & 16.67                                 & 15.27                                 & 11.39                                & 17.53                                 & 15.35                                & 11.87                                & 16.38                                 & 14.98                                 & 11.10                                & 14.50             &      26.28$^{\dagger\dagger}$                \\
Plain Summaries                                  & 14.23                                 & 14.79                                 & 11.13                                & 12.52                                 & 10.91                                & 8.94                                 & 15.90                                 & 14.76                                 & 10.91                                & 12.68       &      15.69$^{\dagger\dagger}$                      \\ 
\bottomrule
\end{tabular}
\caption{Automated evaluation, including {\textbf{\color[HTML]{485aa3}single-LLM prompting}},  {\color[HTML]{d17630}\textbf{fine-tuning}}, {\color[HTML]{48968B}\textbf{multi-LLM prompting}}, and {\color[HTML]{754494}\textbf{our JRE-L framework}}. The option `OS' and `FT' denote `one shot' and `fine-tuning.' Symbols $\dagger$ and $\dagger\dagger$ denote that the statistical significance of the comparison with JRE-L is $p<0.05$ and $p<0.01$, respectively.}
\label{tab:auto}
\end{table*}

\noindent \textbf{Methods For Comparison.} We study models with different sizes, a number of one-shot demonstrations and fine-tuning, as well as previous ASJ baselines and other LLM-collaboration frameworks.
These methods are listed in Appendix \ref{sec:comp} for brevity.
% \begin{itemize}[noitemsep, topsep=1pt]
%     \item \textbf{BART.} \citet{goldsack2022making, cardenas2023don} used BART \citep{LewisBart} for ASJ, showing strong performance.
%     \item \textbf{LLMs.} We test the performance of various LLMs, including both open-source and closed LLMs, i.e., LLaMA-2-7B \citep{touvron2023llama}, Gemma-7B \citep{gemma}, Mistral (7B, 8x7B, \citealp{jiang2023mistral}), Qwen1.5 (7B, 72B, \citealp{bai2023qwen}), LLaMA-3-8B \citep{llama3}, GPT-3.5-Turbo-1106 \citep{openaichatgpt}, and GPT-4-1106-preview \citep{openai2024gpt4}. We prompt these LLMs and also fine-tune the Qwen1.5-7B model.
%     \item \textbf{Our Methods.} We test two versions of collaborating LLMs. One is 2$\times$Qwen1.5-7B+Qwen1.5-1.8B (JRE). Another is this combination where the journalist is replaced by a fine-tuned version as a warm start ($\text{JRE}_{FT}$).
% \end{itemize}

\noindent \textbf{Automatic Evaluation.}
Following \citet{goldsack2022making, cardenas2023don}, we use the Coleman-Liau Index (CLI, \citet{coleman1975computer}), the Flesch-Kincaid Grade Level (FKGL, \citet{kincaid1975derivation}) and the Dale-Chall Readability Score (DCRS, \citet{dale1948formula}) to automatically assess readability.
CLI considers the number of sentences, words, and characters, whereas FKGL is based on the number of sentences, words, and syllables.
DCRS analyzes the average sentence length and the presence of familiar words from a list of the most commonly used words\footnote{\url{https://help.readable.com/en/article/dale-chall-words-list-w877fe}}.

%%%%%%%%%%%%%%%%%%%%%%%%%%%%%%%%%%%%%%%%%%%%%%%%%%%%%%%%%%%%%%%
\begin{table}[t]
\centering
\small
\begin{tabular}{lllll}
\toprule
\textbf{Approach} & \textbf{Read.} & \textbf{Info.} & \textbf{Auth.} & \textbf{Intr.} \\ \hline
\multicolumn{5}{c}{\textbf{Within Field}}                                                                              \\ \hline
Plain   Summaries & 2.95$^{\dagger\dagger}$           & 2.90$^{\dagger\dagger}$           & 3.35$^{\dagger\dagger}$           & 2.70$^{\dagger\dagger}$            \\
Qwen1.5-7B      & 3.50$^\dagger$          & 3.35$^\dagger$           & 3.40$^\dagger$           & 3.10$^\dagger$           \\
GPT-4           & 3.80          & \textbf{3.75} & \textbf{3.80} & 3.40          \\
JRE-L        & \textbf{3.95} & 3.60          & 3.70          & \textbf{3.55} \\
\hline
\multicolumn{5}{c}{\textbf{Outside Field}}                                                 \\ \hline
Plain   Summaries & 2.75$^{\dagger\dagger}$          & 2.85$^{\dagger\dagger}$          & 3.25$^{\dagger\dagger}$          & 2.65$^{\dagger\dagger}$           \\
Qwen1.5-7B      & 3.35$^\dagger$           & 3.10$^\dagger$           & 3.30$^\dagger$           & 3.10$^\dagger$           \\
GPT-4           & 3.40          & \textbf{3.55} & \textbf{3.70} & 3.15          \\
JRE-L        & \textbf{3.65} & 3.40          & 3.55          & \textbf{3.20}  \\ \bottomrule
\end{tabular}
% \caption{Results of human evaluation, where `Read.' indicates `Readability,' `Info.' denotes `Information Conveyance,' `Auth.' represents `Authenticity,' and `Intr.' signifies `Interestingness'.}
\caption{Results of human evaluation. Symbols ``$\dagger$'' and ``$\dagger\dagger$'' indicate that the statistical significance of the comparison with JRE-L is $p<0.05$ and $p<0.01$. The higher the scores, the better.}
\label{tab:human}
\end{table}

\begin{figure*}[ht]
  \centering
  \begin{subfigure}{0.328\textwidth}
    \centering
    \includegraphics[height=0.7\columnwidth]{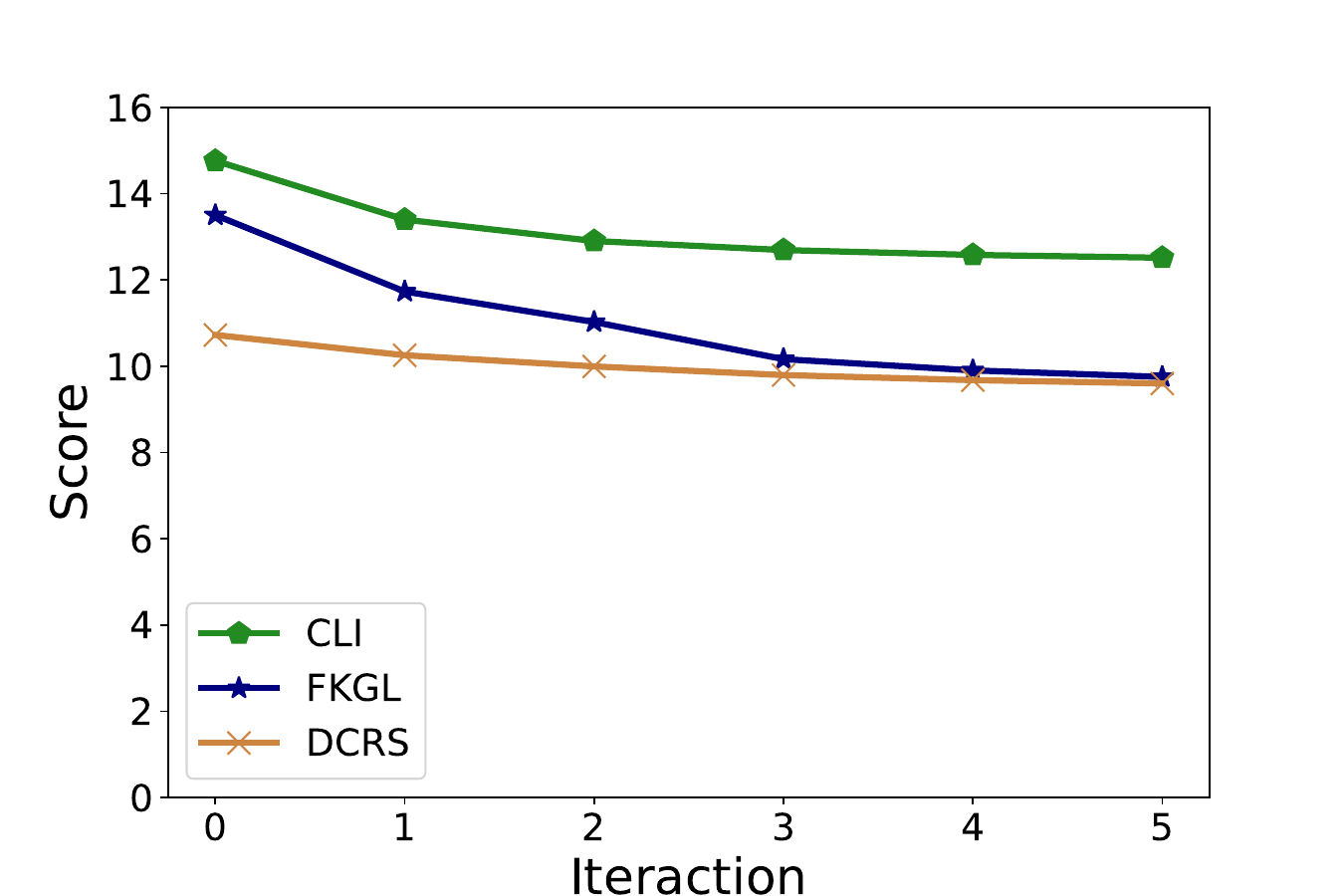}
    \caption{SCITech}
  \end{subfigure}
  \hfill
  \begin{subfigure}{0.329\textwidth}
    \centering
    \includegraphics[height=0.7\columnwidth]{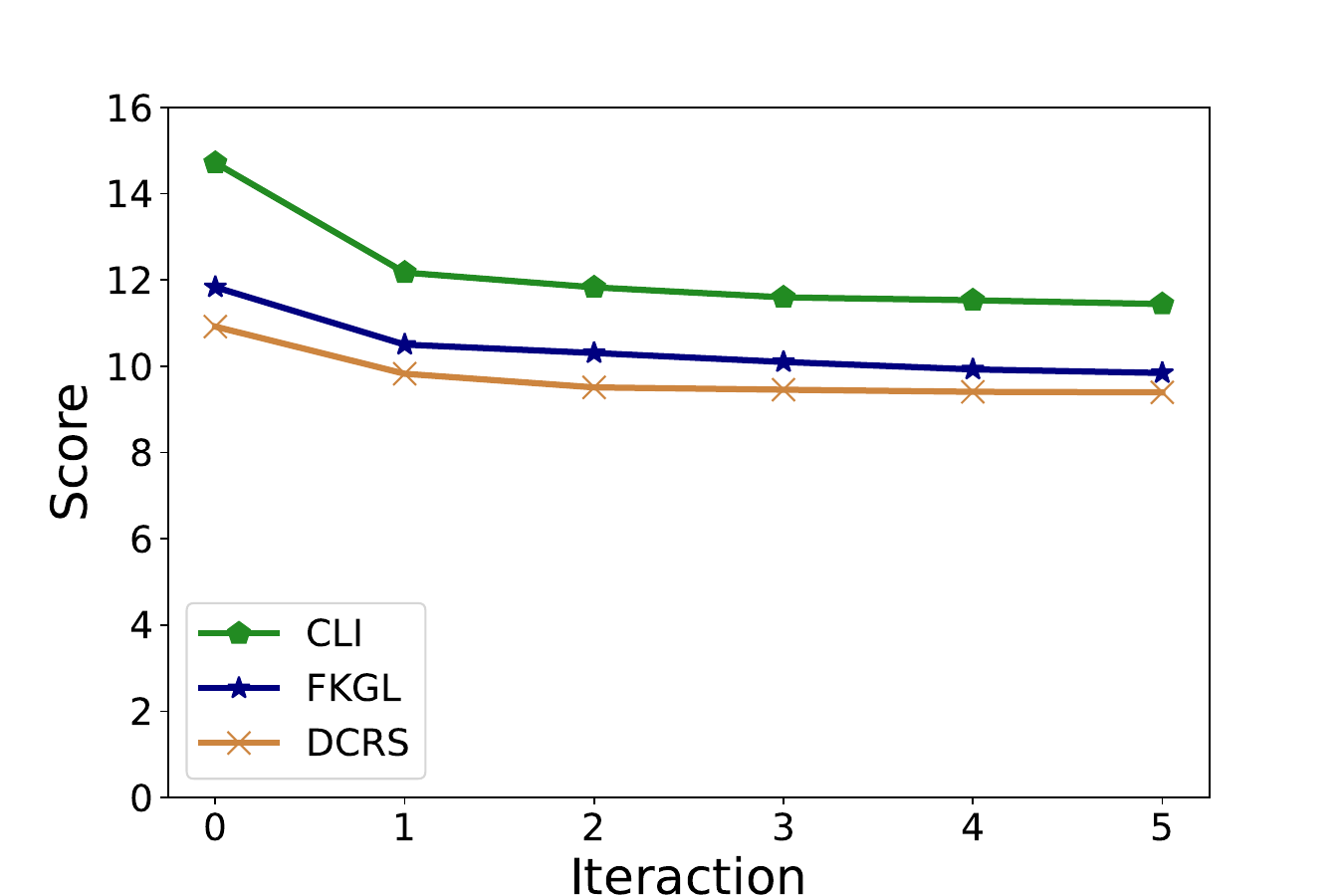}
    \caption{eLife}
  \end{subfigure}
  \hfill
  \begin{subfigure}{0.328\textwidth}
    \centering
    \includegraphics[height=0.7\columnwidth]{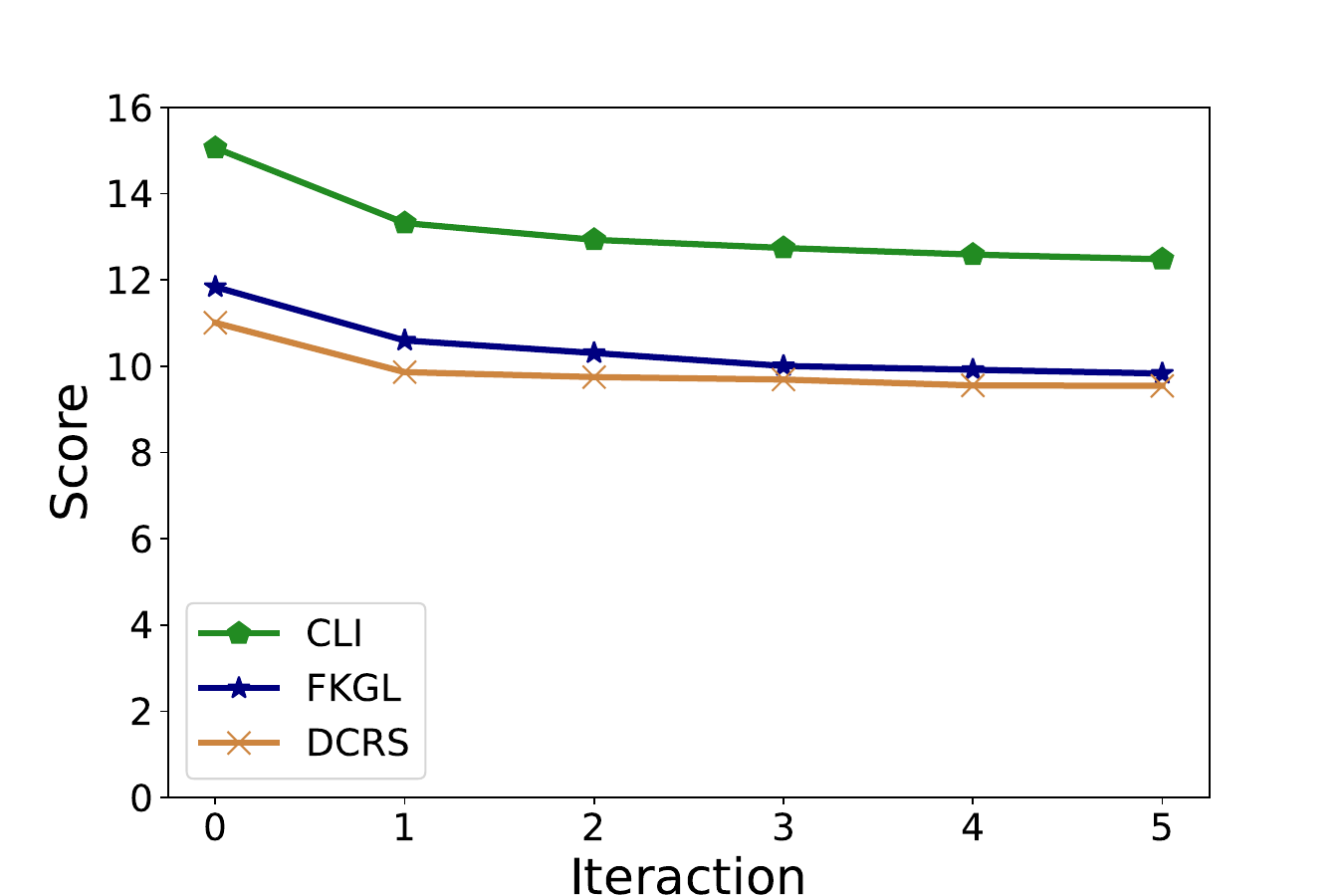}
    \caption{PLOS}
  \end{subfigure}
  \caption{Performance improvement over iterations, with iteration 0 producing the initial writing.}
  \label{fig:trend}
\end{figure*}

%%%%%%%%%%%%%%%%%%%%%%%%%%%%%%%%%%%

\noindent \textbf{Human Evaluation.}
% Automatically assessing the authenticity and informativeness of content has been a challenging task.
% \citet{cardenas2023don} used QuestEval \citep{Scialom_Dray_Lamprier_Piwowarski_Staiano_Wang_Gallinari_2021} to assess the faithfulness of ASJ-generated content, yet the results exhibited significant variances.
% Therefore, human evaluation remains the main method for such assessments.
Four human participants are enlisted for evaluation. 
All of them are pursuing or possess master's degrees, two from computer science and two from biomedical science. 
We sample 10 pairs of original papers in computer science from SCITech and their corresponding popular science articles, as well as 10 pairs in biomedical science, 5 pairs from eLife and 5 pairs from PLOS.
This quantity is comparable to previous studies \citep{goldsack2022making, cardenas2023don}, taking into account both the reliability of the results and the workload of the annotators.

Four representative methods are chosen for human evaluation: (1) plain summaries by human writers, (2) Qwen1.5-7B generation, (3) GPT-4 generation, and (4) generation by our JRE-L.
The human evaluation encompasses multiple dimensions, namely Readability (Read.), Information Conveyance (Info.), Authenticity (Auth.), and Interestingness (Intr.).
Participants are tasked with evaluating the articles using a 1-5 Likert scale \citep{likert1932technique}, grounded on specific questions.
Each participant is assigned to assess all articles both in the field they are familiar with (Within Field) and those they are not familiar with (Outside Field), to provide a genuine evaluation from readers within the specific discipline and general readers.
Appendix \ref{sec:questionnaire} shows the details of these measures and the questionnaire form.

\noindent \textbf{Hyperparameters.}
We list hyperparameters in Appendix \ref{sec:hyper} for brevity.

\begin{table}[t]
    \small
    \centering
    \setlength{\tabcolsep}{5pt}
    \begin{tabular}{lcccc}
    \toprule
    \textbf{Approach} & \textbf{SCITech}$\downarrow$   & \textbf{eLife}$\downarrow$   & \textbf{PLOS}$\downarrow$   & \textbf{Avg.} \\
    \midrule
    JRE-L & \textbf{10.88} & \underline{10.39} & 10.81 & \textbf{10.69} \\
    Reader$\rightarrow$7B  &  \underline{10.95}    &   10.44  &  \textbf{10.79}   & 10.73 \\
    LLM$\rightarrow$LLaMA & 10.99  &  \textbf{10.37}  &   \underline{10.81}  &  \underline{10.72}  \\
    $-$ Reading	& 11.39 & 11.00 &  11.40   & 11.26 \\
    $-$ Suggestions	& 11.44 & 11.03 & 11.49  & 11.32 \\
    $-$ Collaboration	&  11.74  &  11.29  &  11.75  & 11.59 \\
    \bottomrule
    \end{tabular}
    \caption{Ablation results, reporting average scores. The scores in bold are the best, and the underlined scores are the second best.}
    \label{tab:abl_brief}
\end{table}
%%%%%%%%%%%%%%%%%%%%%%%%%%%%%%%%%%%%%%%%%%%%%%%%%%%%%%%%%%%%%%%

\subsection{Automatic Evaluation}
\noindent \textbf{Different Models and Sizes.} Table \ref{tab:auto} show the comparison of different methods.
Recent LLMs of similar scales have shown comparable performance, surpassing LLaMA-2. 
Larger models such as Mixtral-8x7B and Qwen1.5-72B show even better performance, indicating that performance improves as the model scale increases. 
Additionally, the formidable LLM GPT-4 outperforms all the other single LLMs. 
% These findings demonstrate their performance on ASJ consistent with the ability of LLMs.

\noindent \textbf{One-shot.} Among the methods of single-LLM prompting, one-shot demonstration brings some improvements of the performance (Avg.) by 0.32 in Qwen and 0.22 in LLaMA.
In our LLM framework, adding one-shot examples as a warm start shows close performance to the original version.
A possible reason is that these examples can be used as references for the first revision and the iterative process narrows the performance gap.
% However, these references are still technical and cannot provide an effective demonstration for LLM to rewrite papers in a highly readable way.

\noindent \textbf{Fine-tuning.} The fine-tuning methods exhibit competitive performance, slightly better than prompting the open-source LLMs.
Fine-tuning models with a larger scale (Qwen1.5-7B and LLaMA-3-8B) outperforms BART.
Interestingly, in our JRE-L framework, fine-tuning the journalist LLM to warm up does not lead to a noticeable improvement, possibly because the LLM does not follow instructions as well as before after such specialized fine-tuning, thereby affecting cooperation among LLMs.

\noindent \textbf{Previous baselines.} Previous ASJ work \citep{goldsack2022making, cardenas2023don} took BART as their baseline.
Compared to their fine-tuning method, our framework brings an average of $(12.43 - 10.69) / 12.43 = 14\%$ improvement on all the three datasets. 
Moreover, compared with lay summaries written by humans, our method shows higher readability, as those summaries were crafted for technical audiences.

\noindent \textbf{LLM-collaboration frameworks.} 
We adapt two other LLM-collaboration methods to the ASJ task for comparison.
Our methods attain average enhancements of 8.16\% and 10.02\% over them respectively. 
This performance improvement is due to the reader's feedback in our framework, which can better expose writing issues and is not available in the other two paradigms.
This experimental result also demonstrates the effectiveness of our proposed iterative workflow.

% Our collaboration with LLMs has demonstrated significant improvement over previous methods.

% Interestingly, it can be observed that fine-tuning the journalist LLM to warm up does not lead to any noticeable improvement. 
% One possible reason may be the decay in the ability to follow instructions after this specialized training.
% Nonetheless, these empirical results show the effectiveness of our framework that integrates multiple LLMs.

\begin{table*}[t]
\centering
\small
\begin{tabular}{p{15.5cm}}
\toprule
\textbf{Scientific Paper.} In infectious disease diagnosis, \textit{\textbf{results need to be communicated rapidly to healthcare professionals once testing has been completed so that care pathways can be implemented}}. This represents a particular challenge when testing in remote, low-resource rural communities, in which such diseases often create the largest burden. Here, we report a \textit{\textbf{smartphone-based end-to-end platform for multiplexed DNA diagnosis of malaria}}. The approach uses a low-cost paper-based microfluidic diagnostic test, which is combined with deep learning algorithms for local decision support and blockchain technology for secure data connectivity and management. We validated the approach via field tests in rural Uganda, where it correctly identified more than 98\% of tested cases. Our platform also provides secure geotagged diagnostic information, which creates the possibility of integrating infectious disease data within surveillance frameworks.
\\
\midrule
% \textbf{Plain Summary.} Researchers at the U.K.\'s University of Glasgow and Uganda\'s Ministry of Health have developed easy-to-administer blood tests that could help curtail the spread of malaria in sub-Saharan Africa. These "origami" tests can be conducted in the field, with a blood sample taken via fingerprick placed in a channel made from sheets of folded wax paper. The blood is directed into three small chambers of a loop-mediated isothermal amplification (LAMP) machine, which tests for the presence of the mosquito-borne parasite that causes malaria. A smartphone app uses deep learning to ensure the tests are accurate and to monitor community transmission. The results are stored on a blockchain-based ledger to maintain privacy, with anonymized data shared with local authorities. In Uganda, the blood tests were shown to be 98\% accurate.\\
% \hline
\textbf{Article Written by JRE-L.} In the fight against malaria, \vivid{timely diagnosis is crucial}, especially in rural communities where these diseases often hit hardest. Researchers have come up with a brilliant solution using smartphones! It's a \technical{\underline{smartphone-} \underline{based system that combines a low-cost test with deep learning and blockchain technology}}.

First, \vivid{imagine a tiny paper device that can detect malaria DNA right from blood samples}. \technical{\underline{This microfluidic test is as} \underline{simple as using a piece of paper - no fancy lab equipment needed!}} Then, the results are analyzed by smart algorithms on your phone. \vivid{These "AI helpers" give instant feedback to healthcare workers without delay.}

All this diagnostic data is stored securely on blockchain. This means information is protected and cannot be tampered with. In Uganda, field tests showed that this system was almost perfect - it correctly identified 98\% of cases!

\vivid{The best part? The platform also records each diagnosis's location through geotags.} This opens up the possibility to \technical{\underline{connect local disease data to bigger surveillance networks, helping us track and combat malaria more efficiently.}}

So next time you hear about remote villages facing health challenges due to lack of fast diagnosis tools, remember: there's hope with these smartphone-powered innovations!
\\
\bottomrule
\end{tabular}
\caption{An example of the original paper and the article generated by our method. We highlight statements that we regard \vivid{easy to understand} and \technical{\underline{explanations for technical details}}.} 
\label{tab:case}
\end{table*}

%%%%%%%%%%%%%%%%%%%%%%%%%%%%%%%
\begin{table*}[t]
\centering
\small
\begin{tabular}{p{15.5cm}}
\toprule
\textbf{1st Iteration.}
1. The mobile lab on your phone is a portable, paper-based test that uses deep learning algorithms to analyze blood samples for the presence of malaria.
2. The end-to-end system combines a smartphone app, allowing healthcare workers to make informed decisions about treatment without delay.
3. The smartphone app uses \note{deep learning algorithms to interpret the test results}, providing accurate diagnoses over 98\% in field tests.
4. The platform ensures secure data management by using \note{blockchain technology, which records transparently and cannot be tampered with}.
\\
\midrule
\textbf{2nd Iteration.}
1. Smartphones and Blockchain: The article introduces the use of smartphones and \note{blockchain technology in diagnosing diseases like malaria, particularly in remote areas where traditional diagnostic methods may not be available or accessible}.
2. Malaria Diagnosis: The low-cost paper-based microfluidic diagnostic test uses a piece of tiny paper device to detect malaria DNA. It allows rapid detection with the support of AI systems.
3. Low-Cost Test: The AI-based test is referred to as a "smart decision support" system, which uses advanced technology such as artificial intelligence to interpret the results right on the user's device.
4. \note{Artificial Intelligence (AI): The AI acts like a smart detective, quickly identifying potential cases with an accuracy rate of 98\% in real-world tests}, enabling healthcare workers to make informed decisions about treatment without needing fancy labs or waiting too long.
\\
\midrule
\textbf{3rd Iteration.}
1. Smartphone: A device with advanced features such as cameras, sensors, and processors that can perform various tasks, including medical diagnosis.
2. Blood test: A diagnostic procedure where blood is collected from a finger or other body part through a small needle and analyzed for specific signs of malaria (e.g., anemia or fever).
3. \note{AI as personal health assistant: An artificial intelligence system designed to assist individuals in managing their health by providing personalized recommendations based on medical data analysis (e.g., detecting early signs of disease).}
4. Secure digital diary: A digital record containing sensitive information about an individual's health status stored on a secure blockchain network to ensure accuracy and prevent unauthorized access or manipulation.
5. \note{Blockchain technology: A distributed ledger system that allows secure sharing of data across multiple parties without the need for intermediaries or centralized authorities (e.g., storing patient records in hospitals). In this case, it serves as the secure digital ledger for tracking where diseases are spreading in remote areas due to malaria prevalence.}
\\
\bottomrule
\end{tabular}
\caption{Notes taken by the reader, \note{two technical terms and their explanations} are highlighted. The notes become more detailed and comprehensive through the iterative process.} 
\label{tab:notes}
\end{table*}

\subsection{Human Evaluation}
% For a thorough evaluation, we carry out a human assessment on four representative methods. 
Our human participants assessed articles relevant to their fields as well as those in their unfamiliar fields.
The results are reported in Table \ref{tab:human}, where the scores are obtained by averaging scores on the corresponding metric and are categorized into within-field and outside-field articles.
The Krippendorff's alpha value \citep{krippendorff2011computing} between annotators is 0.52, slightly lower than the 0.57 value in \citet{cardenas2023don}, indicating an acceptable agreement.
% \gongyao{
% Notably, there are some differences between the results of the human evaluation and the automatic evaluation, and both have their advantages and disadvantages.
% There is some subjectivity in the human evaluation, while the automatic evaluation is based on statistics and cannot cover all cases.
% }
Notably, participants assign lower ratings to articles outside their expertise than those within their expertise areas, indicating the effect of topic familiarity on reader experience.
Nevertheless, the relative performance between methods appears consistent across both within expertise and outside expertise settings.

Interestingly, all LLM-based methods outperform the plain summaries written by humans ($p<0.01$), probably because these summaries are still too technical for readers, leading to lower ratings.
The LLM-generated contents, targeted at general audience, are easier for both in-domain and out-of-domain readers to read.
Furthermore, our JRE-L surpasses the single Qwen in all dimensions ($p<0.05$), demonstrating the effectiveness of LLM collaboration.
Our JRE-L achieves high ratings in all dimensions, close to those of GPT-4, even though our model is much smaller.
Collectively, these findings attest to the effectiveness of our proposed approach.

\section{Analysis}
\label{sec:ana}

\subsection{Performance Over Iterations}
As our system is an iterative enhancement framework, we study its performance over iterations.
% Thus, we conduct an in-depth analysis of the performance change throughout the iterative cycles.
The entire writing-reading-suggestion-revision process is carried out for five iterations, with readability scores depicted in Figure \ref{fig:trend}.
The figure shows a pronounced decline in reading difficulty in the first iteration and the decline continues in the next two iterations.
This pattern demonstrates the efficacy of our iterative revision methodology.
Following the third iteration, the performance levels off, indicating diminishing improvements from subsequent suggestion and editing efforts.

\subsection{Ablation Study}
We conduct an ablation experiment to evaluate the effectiveness of each component by removing or replacing components.
Table \ref{tab:abl_brief} presents the averaged results on each dataset, and detailed results are included in Appendix \ref{sec:abl_all}.
In the ``\textit{Reader$\rightarrow$7B}'' setting, we substitute the 1.8B reader model with the 7B version. 
This substitution leads to a minor performance fluctuation.
On one hand, the 7B model tolerates low readability of content and highlights fewer writing issues.
On the other hand, it excels in instruction following, enhancing task execution and reducing intermediate errors in the workflow. 
The dynamics between gains and losses render the 7B reader comparably advantageous at times and disadvantageous at others.
As such, we recommend using the 1.8B model for its higher resource efficiency and throughput.
To test the generalizability of our framework, we replace each agent with LLaMA models of 3-8B in the “\textit{LLM$\rightarrow$LLaMA}” setup. 
Experimental results show that the performance after replacement is close to that based on Qwen, demonstrating that our framework can be adapted to different LLMs.

In $-\textit{Reading Notes}$, we eliminate the requirement for the reader LLM to read the article and make notes. Instead, we have the editor LLM offer suggestions directly, matching the situation in Figure \ref{fig:read} (a).
In $-$\textit{Suggestions}, the editor is omitted, and the journalist revises the article based on the reader LLM's reading.
In $-$\textit{Collaboration}, the journalist revises the article based on the previous writing without any input from the reader or the editor.
As depicted in Table \ref{tab:stat}, our approach exhibits a decrease in performance when each module and the collaboration is removed, underscoring the significance of each module and the collaboration.

\subsection{Writing Case Analysis}

To facilitate an intuitive assessment of our method, we present a case study on one writing sample. 
As shown in Table \ref{tab:case}, our method can generate articles that are more readable, with concise and vivid narratives, along with explanations for technical details.
For instance, our generated article states that ``timely diagnosis is crucial'' rather than ``results need to be communicated rapidly to healthcare professionals once testing has been completed so that care pathways can be implemented'', making it more brief and accessible for readers to grasp the research objective.
Moreover, our generated article details that the proposed system is a ``smartphone-based system that combines a low-cost test with deep learning and blockchain technology'' rather than ``smartphone-based end-to-end platform for multiplexed DNA diagnosis of malaria'', enhancing the comprehensibility for a broader audience.

\subsection{Case of Reading Notes}
\label{sec:stu}

% We further present a case study on the notes taken by the reader LLM during the first three iterations.\footnote{We also examine the suggestions given by the editor LLM for the revision over the iterative process in Appendix \ref{sec:suggestion}.}
We further present a case study on the notes taken by the reader LLM during the first three iterations.
As shown in Table \ref{tab:notes}, detailed explanations for technical terms increase with iterations.
For example, the technical terms ``AI'' and ``blockchain'' are more thoroughly explained in the third iteration than in the first two iterations. 
This phenomenon suggests that as the readability of writing improves, readers can understand the content better and provide more elaborate explanations, aligning with human reading behaviors.
% We also examine the suggestions given by the editor LLM for the revision over the iterative process in Appendix \ref{sec:suggestion}.

\subsection{Suggestions for Revision}
\label{sec:suggestion}
We further examine the suggestions given by the editor LLM for the revision over the iterative process, as shown in Table \ref{tab:suggest}.
In the initial stage, the editor offers general advice. 
Following the revision, the advice becomes more specific and detailed. 
Subsequent rounds of suggestions highlight specific content in the article and recommend revisions in various aspects. 
In the 3rd iteration, the feedback from the editor suggests on phrases in the previous version.
This revision behaviour is similar to real-world science journalism, showcasing the efficacy of our framework and the social intelligence of LLMs.

%%%%%%%%%%%%%%%%%%%%%%%%%%%%%%
\begin{table*}[t]
\centering
\small
\begin{tabular}{p{15.4cm}}
\toprule
\textbf{1st Iteration.}
1. Simplify technical terms
2. Break down processes
3. Emphasize benefits
4. Conclusion statement
\\
\midrule
\textbf{2nd Iteration.}
1. \textbf{Emphasize technology}: Instead of listing specific platforms (Ethereum or Hyperledger Fabric), mention that a secure digital ledger is used without specifying which one.
2. \textbf{Simplify medical terms}: For "Plasmodium falciparum," provide a simpler explanation or use a common term like "malaria parasite type."
3. \textbf{Break down process}: Explain how smartphone analysis works in more detail - perhaps by describing how it compares to traditional lab methods.
4. \textbf{Connect to equity}: Highlight how this technology addresses health disparities by providing quick diagnosis in remote areas.
\\
\midrule
\textbf{3rd Iteration.}
1. \textbf{Emphasize simplicity}: For accessibility, rephrase "low-cost paper-based microfluidic diagnostic test" as "affordable, easy-to-use test with a paper strip."
2. \textbf{Explain AI in simpler terms}: Instead of "AI instantly interprets results," say "The smartphone app quickly analyzes the data to give a diagnosis."
3. \textbf{Break down data security}: Highlight that information is stored securely on a phone or cloud server with strong passwords or encryption.
4. \textbf{Quantify success}: Mention that 98\% accuracy rate is exceptional but could be framed as an impressive achievement ("This system detected almost all cases correctly!").
5. \textbf{Cite real-life impact}: Share examples of how this technology has made a difference in remote communities to connect it emotionally with readers.
\\
\bottomrule
\end{tabular}
\caption{Suggestions provided by the editor LLM are becoming increasingly specific through the iterative process.} 
\label{tab:suggest}
\end{table*}

% \subsection{Suggestions for Revision}
% We further examine the advice given by the LLM editor for the revision over the iterative process, as exemplified in Table \ref{tab:suggest}.
% In the initial stage, the editor offers general advice. 
% Following the revision, the advice becomes more specific and detailed. 
% Subsequent rounds of suggestions highlight specific content in the article and recommend revisions based on various aspects. 
% In the 3rd iteration, the feedback from the editor becomes more detailed and specific, with specific adjustments from the original version to the revised content.
% This revision behaviour is similar to real-world science journalism, showcasing the efficacy of our framework and the social intelligence of LLMs.
\section{Conclusion}
This study proposes the collaboration of LLMs in the loop for ASJ aimed at general readers.
Initially, an LLM functions as a journalist by composing an explanatory article for the general public. 
Subsequently, another LLM, acting as a general audience, reads these articles and takes notes, helping reveal the readability issues in the generated article.
Then, an editor LLM assesses the reader's notes and offers suggestions for improvement. 
Following the suggestions, the journalist revises its article, and then passes it to the reader to continue the iterative process.
Extensive experiments are conducted to evaluate the effectiveness of our framework, including both automatic and human evaluation. 
In comparison to prompting and fine-tuning LLMs as other ASJ systems do, our method achieves the highest readability while maintaining high quality.
% Additionally, an in-depth analysis further examines and discusses such LLM-driven ASJ systems.

\section*{Limitations and Future Work}
We identify the following five limitations of our work.
First, following previous work, we have defined ASJ as the process of transforming a single paper into an article intended for a general audience. 
In practice, a popular science article may encompass multiple studies. 
Therefore, an extension can be the consolidation of several papers into a single article.
Second, we have utilized some statistical indexes for automatic assessment, but these statistical measures may miss semantic information. 
LLM-driven evaluation could offer a solution. 
While there remains a gap between LLMs and humans in evaluating text on readability and authenticity, efforts such as human-preference optimization could be made to minimize this gap.
Third, given our exploratory approach in utilizing LLMs for ASJ, we strategically chose abstracts as opposed to full papers as input to maintain both simplicity and resource efficiency.
Nevertheless, long-context ASJ is an intriguing task with higher impact.
Fourth, due to the limit of space and effort, we study our framework on settings with up to three LLMs.
It could be interesting to study collaborative writing between writers, receiving feedback from multiple readers of different backgrounds, and considering revision suggestions from a hierarchy of editors or editors from different areas.
Lastly, all components of our framework are powered by LLMs. 
In addition to our efforts to make each LLM simulate humans, it will be interesting to incorporate genuine human preferences to enhance the generated content.

\section*{Ethics Statement}
The experiments in this study were conducted on publicly available datasets. 
Any information involving privacy was removed. 
All annotators have been properly paid for their efforts.
% Each participant received $\$$70 for completing the human evaluation task, a compensation well above the local income levels.

\section*{Acknowledgements}
We would like to thank the anonymous reviewers for their constructive comments, which help improve this work.
This research was partially supported by the Guangzhou Municipality Big Data Intelligence Key Lab (2023A03J0012).

% \section*{Acknowledgments}

%\bibliography{anthology, ,custom}
\bibliography{custom}

\begin{thebibliography}{60}
\providecommand{\natexlab}[1]{#1}

\bibitem[{Achiam et~al.(2023)Achiam, Adler, Agarwal, Ahmad, Akkaya, Aleman, Almeida, Altenschmidt, and et~al.}]{openai2024gpt4}
Josh Achiam, Steven Adler, Sandhini Agarwal, Lama Ahmad, Ilge Akkaya, Florencia~Leoni Aleman, Diogo Almeida, Janko Altenschmidt, and Sam~Altman et~al. 2023.
\newblock Gpt-4 technical report.
\newblock \emph{arXiv preprint arXiv:2303.08774}.

\bibitem[{Allan(2011)}]{allan2011introduction}
Stuart Allan. 2011.
\newblock Introduction: Science journalism in a digital age.
\newblock \emph{Journalism}, 12(7):771--777.

\bibitem[{Anderson et~al.(2015)Anderson, Bell, and Shirky}]{anderson2015post}
Christopher~William Anderson, Emily Bell, and Clay Shirky. 2015.
\newblock Post-industrial journalism: Adapting to the present.
\newblock \emph{Geopolitics, History and International Relations}, 7(2):32.

\bibitem[{Angler(2017)}]{angler2017science}
Martin Angler. 2017.
\newblock \emph{Science journalism: an introduction}.
\newblock Routledge.

\bibitem[{August et~al.(2024)August, Lo, Smith, and Reinecke}]{august2024know}
Tal August, Kyle Lo, Noah~A Smith, and Katharina Reinecke. 2024.
\newblock Know your audience: The benefits and pitfalls of generating plain language summaries beyond the" general" audience.
\newblock \emph{arXiv preprint arXiv:2403.04979}.

\bibitem[{Baek et~al.(2024)Baek, Jauhar, Cucerzan, and Hwang}]{baek2024researchagent}
Jinheon Baek, Sujay~Kumar Jauhar, Silviu Cucerzan, and Sung~Ju Hwang. 2024.
\newblock Researchagent: Iterative research idea generation over scientific literature with large language models.
\newblock \emph{arXiv preprint arXiv:2404.07738}.

\bibitem[{Bai et~al.(2023)Bai, Bai, Chu, Cui, Dang, Deng, Fan, Ge, Han, Huang et~al.}]{bai2023qwen}
Jinze Bai, Shuai Bai, Yunfei Chu, Zeyu Cui, Kai Dang, Xiaodong Deng, Yang Fan, Wenbin Ge, Yu~Han, Fei Huang, et~al. 2023.
\newblock Qwen technical report.
\newblock \emph{arXiv preprint arXiv:2309.16609}.

\bibitem[{Brownell et~al.(2013)Brownell, Price, and Steinman}]{brownell2013science}
Sara~E Brownell, Jordan~V Price, and Lawrence Steinman. 2013.
\newblock Science communication to the general public: why we need to teach undergraduate and graduate students this skill as part of their formal scientific training.
\newblock \emph{Journal of undergraduate neuroscience education}, 12(1):E6.

\bibitem[{Bryant(2002)}]{bryant2002fluid}
John Bryant. 2002.
\newblock \emph{The fluid text: A theory of revision and editing for book and screen}.
\newblock University of Michigan Press.

\bibitem[{Cardenas et~al.(2023)Cardenas, Yao, Wang, and Hou}]{cardenas2023don}
Ronald Cardenas, Bingsheng Yao, Dakuo Wang, and Yufang Hou. 2023.
\newblock ‘don’t get too technical with me’: A discourse structure-based framework for automatic science journalism.
\newblock In \emph{Proceedings of the 2023 Conference on Empirical Methods in Natural Language Processing}, pages 1186--1202.

\bibitem[{Caselli et~al.(2015)Caselli, Vossen, van Erp, Fokkens, Ilievski, Izquierdo, Le, Morante, and Postma}]{caselli2015s}
Tommaso Caselli, Piek Vossen, Marieke van Erp, Antske Fokkens, Filip Ilievski, Rub{\'e}n Izquierdo, Minh Le, Roser Morante, and Marten Postma. 2015.
\newblock When it's all piling up: investigating error propagation in an nlp pipeline.
\newblock In \emph{WNACP@ NLDB}.

\bibitem[{Chan et~al.(2023)Chan, Chen, Su, Yu, Xue, Zhang, Fu, and Liu}]{chan2023chateval}
Chi-Min Chan, Weize Chen, Yusheng Su, Jianxuan Yu, Wei Xue, Shanghang Zhang, Jie Fu, and Zhiyuan Liu. 2023.
\newblock Chateval: Towards better llm-based evaluators through multi-agent debate.
\newblock \emph{arXiv preprint arXiv:2308.07201}.

\bibitem[{Chen et~al.(2023)Chen, Wang, Huo, Li, and Zhang}]{chen2023gamegpt}
Dake Chen, Hanbin Wang, Yunhao Huo, Yuzhao Li, and Haoyang Zhang. 2023.
\newblock Gamegpt: Multi-agent collaborative framework for game development.
\newblock \emph{arXiv preprint arXiv:2310.08067}.

\bibitem[{Chen et~al.(2024)Chen, Li, and Niu}]{chenboosting2024}
Sijia Chen, Baochun Li, and Di~Niu. 2024.
\newblock Boosting of thoughts: Trial-and-error problem solving with large language models.
\newblock In \emph{The Twelfth International Conference on Learning Representations}.

\bibitem[{Cho and MacArthur(2011)}]{cho2011learning}
Kwangsu Cho and Charles MacArthur. 2011.
\newblock Learning by reviewing.
\newblock \emph{Journal of educational psychology}, 103(1):73.

\bibitem[{Clark et~al.(2016)Clark, Russell, Enyeart, Gracia, Wessel, Jarmoskaite, Polioudakis, Stuart, Gonzalez, MacKrell et~al.}]{clark2016science}
Greg Clark, Josh Russell, Peter Enyeart, Brant Gracia, Aimee Wessel, Inga Jarmoskaite, Damon Polioudakis, Yoel Stuart, Tony Gonzalez, Al~MacKrell, et~al. 2016.
\newblock Science educational outreach programs that benefit students and scientists.
\newblock \emph{PLoS Biology}, 14(2):e1002368.

\bibitem[{Coleman and Liau(1975)}]{coleman1975computer}
Meri Coleman and Ta~Lin Liau. 1975.
\newblock A computer readability formula designed for machine scoring.
\newblock \emph{Journal of Applied Psychology}, 60(2):283.

\bibitem[{Dale and Chall(1948)}]{dale1948formula}
Edgar Dale and Jeanne~S Chall. 1948.
\newblock A formula for predicting readability: Instructions.
\newblock \emph{Educational research bulletin}, pages 37--54.

\bibitem[{Dangovski et~al.(2021)Dangovski, Shen, Byrd, Jing, Tsvetkova, Nakov, and Solja{\v{c}}i{\'c}}]{dangovski2021we}
Rumen Dangovski, Michelle Shen, Dawson Byrd, Li~Jing, Desislava Tsvetkova, Preslav Nakov, and Marin Solja{\v{c}}i{\'c}. 2021.
\newblock We can explain your research in layman's terms: Towards automating science journalism at scale.
\newblock In \emph{Proceedings of the AAAI Conference on Artificial Intelligence}, pages 12728--12737.

\bibitem[{Desmond et~al.(2024)Desmond, Ashktorab, Pan, Dugan, and Johnson}]{desmond2024evalullm}
Michael Desmond, Zahra Ashktorab, Qian Pan, Casey Dugan, and James~M Johnson. 2024.
\newblock Evalullm: Llm assisted evaluation of generative outputs.
\newblock In \emph{Companion Proceedings of the 29th International Conference on Intelligent User Interfaces}, pages 30--32.

\bibitem[{Ding et~al.(2023)Ding, Chen, Fang, Liu, Qiu, and Chai}]{ding2023designgpt}
Shiying Ding, Xinyi Chen, Yan Fang, Wenrui Liu, Yiwu Qiu, and Chunlei Chai. 2023.
\newblock Designgpt: Multi-agent collaboration in design.
\newblock In \emph{2023 16th International Symposium on Computational Intelligence and Design (ISCID)}, pages 204--208. IEEE.

\bibitem[{Dziri et~al.(2023)Dziri, Lu, Sclar, Li, Jiang, Lin, Welleck, West, Bhagavatula, Bras, Hwang, Sanyal, Ren, Ettinger, Harchaoui, and Choi}]{dziri2024faith}
Nouha Dziri, Ximing Lu, Melanie Sclar, Xiang~Lorraine Li, Liwei Jiang, Bill~Yuchen Lin, Sean Welleck, Peter West, Chandra Bhagavatula, Ronan~Le Bras, Jena~D. Hwang, Soumya Sanyal, Xiang Ren, Allyson Ettinger, Za{\"{\i}}d Harchaoui, and Yejin Choi. 2023.
\newblock \href {http://papers.nips.cc/paper\_files/paper/2023/hash/deb3c28192f979302c157cb653c15e90-Abstract-Conference.html} {Faith and fate: Limits of transformers on compositionality}.
\newblock In \emph{Advances in Neural Information Processing Systems 36: Annual Conference on Neural Information Processing Systems 2023, NeurIPS 2023, New Orleans, LA, USA, December 10 - 16, 2023}.

\bibitem[{Goldsack et~al.(2022)Goldsack, Zhang, Lin, and Scarton}]{goldsack2022making}
Tomas Goldsack, Zhihao Zhang, Chenghua Lin, and Carolina Scarton. 2022.
\newblock Making science simple: Corpora for the lay summarisation of scientific literature.
\newblock In \emph{Proceedings of the 2022 Conference on Empirical Methods in Natural Language Processing}, pages 10589--10604.

\bibitem[{G{\"o}pfert(2008)}]{gopfert2008strength}
Winfried G{\"o}pfert. 2008.
\newblock The strength of pr and the weakness of science journalism.
\newblock In \emph{Journalism, science and society}, pages 227--238. Routledge.

\bibitem[{Hu et~al.(2021)Hu, Shen, Wallis, Allen-Zhu, Li, Wang, Wang, and Chen}]{hu2021lora}
Edward~J Hu, Yelong Shen, Phillip Wallis, Zeyuan Allen-Zhu, Yuanzhi Li, Shean Wang, Lu~Wang, and Weizhu Chen. 2021.
\newblock Lora: Low-rank adaptation of large language models.
\newblock \emph{arXiv preprint arXiv:2106.09685}.

\bibitem[{Jiang et~al.(2023)Jiang, Sablayrolles, Mensch, Bamford, Chaplot, Casas, Bressand, Lengyel, Lample, Saulnier et~al.}]{jiang2023mistral}
Albert~Q Jiang, Alexandre Sablayrolles, Arthur Mensch, Chris Bamford, Devendra~Singh Chaplot, Diego de~las Casas, Florian Bressand, Gianna Lengyel, Guillaume Lample, Lucile Saulnier, et~al. 2023.
\newblock Mistral 7b.
\newblock \emph{arXiv preprint arXiv:2310.06825}.

\bibitem[{Kincaid et~al.(1975)Kincaid, Fishburne, Rogers, and Chissom}]{kincaid1975derivation}
J~Peter Kincaid, RP~Fishburne, RL~Rogers, and BS~Chissom. 1975.
\newblock Derivation of new readability formulas (automated reliability index, fog count and flesch reading ease formula) for navy enlisted personnel (research branch report 8-75). memphis, tn: Naval air station; 1975.
\newblock \emph{Naval Technical Training, US Naval Air Station: Millington, TN}.

\bibitem[{Krippendorff(2011)}]{krippendorff2011computing}
Klaus Krippendorff. 2011.
\newblock Computing krippendorff’s alpha-reliability.

\bibitem[{Kumar et~al.(2024)Kumar, Kohli, Ghosal, and Ekbal}]{kumar2024longform}
Sandeep Kumar, Guneet~Singh Kohli, Tirthankar Ghosal, and Asif Ekbal. 2024.
\newblock Longform multimodal lay summarization of scientific papers: Towards automatically generating science blogs from research articles.
\newblock In \emph{Proceedings of the 2024 Joint International Conference on Computational Linguistics, Language Resources and Evaluation (LREC-COLING 2024)}, pages 10790--10801.

\bibitem[{Lee et~al.(2024)Lee, Gero, Chung, Shum, Raheja, Shen, Venugopalan, Wambsganss, Zhou, Alghamdi et~al.}]{lee2024design}
Mina Lee, Katy~Ilonka Gero, John Joon~Young Chung, Simon~Buckingham Shum, Vipul Raheja, Hua Shen, Subhashini Venugopalan, Thiemo Wambsganss, David Zhou, Emad~A Alghamdi, et~al. 2024.
\newblock A design space for intelligent and interactive writing assistants.
\newblock In \emph{Proceedings of the CHI Conference on Human Factors in Computing Systems}, pages 1--35.

\bibitem[{Lewis et~al.(2020)Lewis, Liu, Goyal, Ghazvininejad, Mohamed, Levy, Stoyanov, and Zettlemoyer}]{LewisBart}
Mike Lewis, Yinhan Liu, Naman Goyal, Marjan Ghazvininejad, Abdelrahman Mohamed, Omer Levy, Veselin Stoyanov, and Luke Zettlemoyer. 2020.
\newblock \href {https://doi.org/10.18653/v1/2020.acl-main.703} {Bart: Denoising sequence-to-sequence pre-training for natural language generation, translation, and comprehension.}
\newblock In \emph{Proceedings of the 58th Annual Meeting of the Association for Computational Linguistics}.

\bibitem[{Likert(1932)}]{likert1932technique}
Rensis Likert. 1932.
\newblock A technique for the measurement of attitudes.
\newblock \emph{Archives of psychology}.

\bibitem[{Lin et~al.(2023)Lin, Tang, Tang, Yang, Dang, and Han}]{lin2023awq}
Ji~Lin, Jiaming Tang, Haotian Tang, Shang Yang, Xingyu Dang, and Song Han. 2023.
\newblock Awq: Activation-aware weight quantization for llm compression and acceleration.
\newblock \emph{arXiv preprint arXiv:2306.00978}.

\bibitem[{Liu et~al.(2023)Liu, Zhang, Li, Liu, and Yang}]{liu2023dynamic}
Zijun Liu, Yanzhe Zhang, Peng Li, Yang Liu, and Diyi Yang. 2023.
\newblock Dynamic llm-agent network: An llm-agent collaboration framework with agent team optimization.
\newblock \emph{arXiv preprint arXiv:2310.02170}.

\bibitem[{{Meta}(2024)}]{llama3}
{Meta}. 2024.
\newblock \href {https://ai.meta.com/blog/meta-llama-3/} {{Introducing Meta Llama 3: The most capable openly available LLM to date}}.

\bibitem[{Nip(2006)}]{nip2006exploring}
Joyce~YM Nip. 2006.
\newblock Exploring the second phase of public journalism.
\newblock \emph{Journalism studies}, 7(2):212--236.

\bibitem[{{OpenAI}(2023)}]{openaichatgpt}
{OpenAI}. 2023.
\newblock \href {https://openai.com/blog/chatgpt} {{Introducing ChatGPT}}.

\bibitem[{Park et~al.(2023)Park, O'Brien, Cai, Morris, Liang, and Bernstein}]{park2023generative}
Joon~Sung Park, Joseph O'Brien, Carrie~Jun Cai, Meredith~Ringel Morris, Percy Liang, and Michael~S Bernstein. 2023.
\newblock Generative agents: Interactive simulacra of human behavior.
\newblock In \emph{Proceedings of the 36th Annual ACM Symposium on User Interface Software and Technology}, pages 1--22.

\bibitem[{Pu et~al.(2024)Pu, Wang, Loy, and Demberg}]{pu2024scinews}
Dongqi Pu, Yifan Wang, Jia Loy, and Vera Demberg. 2024.
\newblock Scinews: From scholarly complexities to public narratives--a dataset for scientific news report generation.
\newblock \emph{arXiv preprint arXiv:2403.17768}.

\bibitem[{Qian et~al.(2024)Qian, Liu, Liu, Chen, Dang, Li, Yang, Chen, Su, Cong et~al.}]{qian2023communicative}
Chen Qian, Wei Liu, Hongzhang Liu, Nuo Chen, Yufan Dang, Jiahao Li, Cheng Yang, Weize Chen, Yusheng Su, Xin Cong, et~al. 2024.
\newblock Chatdev: Communicative agents for software development.
\newblock In \emph{Proceedings of the 62nd Annual Meeting of the Association for Computational Linguistics (Volume 1: Long Papers)}, pages 15174--15186.

\bibitem[{Qin et~al.(2024)Qin, Liang, Ye, Zhu, Yan, Lu, Lin, Cong, Tang, Qian, Zhao, Hong, Tian, Xie, Zhou, Gerstein, Li, Liu, and Sun}]{qin2024tool}
Yujia Qin, Shihao Liang, Yining Ye, Kunlun Zhu, Lan Yan, Yaxi Lu, Yankai Lin, Xin Cong, Xiangru Tang, Bill Qian, Sihan Zhao, Lauren Hong, Runchu Tian, Ruobing Xie, Jie Zhou, Mark Gerstein, Dahai Li, Zhiyuan Liu, and Maosong Sun. 2024.
\newblock \href {https://openreview.net/forum?id=dHng2O0Jjr} {Toolllm: Facilitating large language models to master 16000+ real-world apis}.
\newblock In \emph{The Twelfth International Conference on Learning Representations, {ICLR} 2024, Vienna, Austria, May 7-11, 2024}. OpenReview.net.

\bibitem[{Scialom et~al.(2021)Scialom, Dray, Lamprier, Piwowarski, Staiano, Wang, and Gallinari}]{Scialom_Dray_Lamprier_Piwowarski_Staiano_Wang_Gallinari_2021}
Thomas Scialom, Paul-Alexis Dray, Sylvain Lamprier, Benjamin Piwowarski, Jacopo Staiano, Alex Wang, and Patrick Gallinari. 2021.
\newblock Questeval: Summarization asks for fact-based evaluation.
\newblock \emph{Empirical Methods in Natural Language Processing}.

\bibitem[{Senel et~al.(2024)Senel, Fetahu, Yoshida, Chen, Castellucci, Vedula, Choi, and Malmasi}]{senel-etal-2024-generative}
L{\"u}tfi~Kerem Senel, Besnik Fetahu, Davis Yoshida, Zhiyu Chen, Giuseppe Castellucci, Nikhita Vedula, Jason~Ingyu Choi, and Shervin Malmasi. 2024.
\newblock \href {https://aclanthology.org/2024.acl-long.295} {Generative explore-exploit: Training-free optimization of generative recommender systems using {LLM} optimizers}.
\newblock In \emph{Proceedings of the 62nd Annual Meeting of the Association for Computational Linguistics (Volume 1: Long Papers)}, pages 5396--5420, Bangkok, Thailand. Association for Computational Linguistics.

\bibitem[{Talebirad and Nadiri(2023)}]{talebirad2023multi}
Yashar Talebirad and Amirhossein Nadiri. 2023.
\newblock Multi-agent collaboration: Harnessing the power of intelligent llm agents.
\newblock \emph{arXiv preprint arXiv:2306.03314}.

\bibitem[{Tkachenko et~al.(2020-2022)Tkachenko, Malyuk, Holmanyuk, and Liubimov}]{Label-Studio}
Maxim Tkachenko, Mikhail Malyuk, Andrey Holmanyuk, and Nikolai Liubimov. 2020-2022.
\newblock \href {https://github.com/heartexlabs/label-studio} {{Label Studio}: Data labeling software}.
\newblock Open source software available from https://github.com/heartexlabs/label-studio.

\bibitem[{Touvron et~al.(2023)Touvron, Martin, Stone, Albert, Almahairi, Babaei, Bashlykov, Batra, Bhargava, Bhosale et~al.}]{touvron2023llama}
Hugo Touvron, Louis Martin, Kevin Stone, Peter Albert, Amjad Almahairi, Yasmine Babaei, Nikolay Bashlykov, Soumya Batra, Prajjwal Bhargava, Shruti Bhosale, et~al. 2023.
\newblock Llama 2: Open foundation and fine-tuned chat models.
\newblock \emph{arXiv preprint arXiv:2307.09288}.

\bibitem[{Venkatraman et~al.(2024)Venkatraman, Tripto, and Lee}]{venkatraman2024collabstory}
Saranya Venkatraman, Nafis~Irtiza Tripto, and Dongwon Lee. 2024.
\newblock Collabstory: Multi-llm collaborative story generation and authorship analysis.
\newblock \emph{arXiv preprint arXiv:2406.12665}.

\bibitem[{Wang et~al.(2024{\natexlab{a}})Wang, Downey, Ji, and Hope}]{wang2024scimon}
Qingyun Wang, Doug Downey, Heng Ji, and Tom Hope. 2024{\natexlab{a}}.
\newblock \href {https://doi.org/10.18653/V1/2024.ACL-LONG.18} {Scimon: Scientific inspiration machines optimized for novelty}.
\newblock In \emph{Proceedings of the 62nd Annual Meeting of the Association for Computational Linguistics (Volume 1: Long Papers), {ACL} 2024, Bangkok, Thailand, August 11-16, 2024}, pages 279--299. Association for Computational Linguistics.

\bibitem[{Wang et~al.(2024{\natexlab{b}})Wang, Guo, Yao, Zhang, Zhang, Wu, Zhang, Dai, Zhang, Wen, Ye, Zhang, and Zhang}]{wang2024auto}
Yidong Wang, Qi~Guo, Wenjin Yao, Hongbo Zhang, Xin Zhang, Zhen Wu, Meishan Zhang, Xinyu Dai, Min Zhang, Qingsong Wen, Wei Ye, Shikun Zhang, and Yue Zhang. 2024{\natexlab{b}}.
\newblock \href {https://doi.org/10.48550/ARXIV.2406.10252} {Autosurvey: Large language models can automatically write surveys}.
\newblock \emph{CoRR}, abs/2406.10252.

\bibitem[{Wasi et~al.(2024)Wasi, Islam, and Islam}]{wasi2024llms}
Azmine~Toushik Wasi, Rafia Islam, and Raima Islam. 2024.
\newblock Llms as writing assistants: Exploring perspectives on sense of ownership and reasoning.
\newblock \emph{arXiv preprint arXiv:2404.00027}.

\bibitem[{Wolf et~al.(2019)Wolf, Debut, Sanh, Chaumond, Delangue, Moi, Cistac, Rault, Louf, Funtowicz et~al.}]{wolf2019huggingface}
Thomas Wolf, Lysandre Debut, Victor Sanh, Julien Chaumond, Clement Delangue, Anthony Moi, Pierric Cistac, Tim Rault, R{\'e}mi Louf, Morgan Funtowicz, et~al. 2019.
\newblock Huggingface's transformers: State-of-the-art natural language processing.
\newblock \emph{arXiv preprint arXiv:1910.03771}.

\bibitem[{Wu et~al.(2018)Wu, Tan, He, Tian, Qin, Lai, and Liu}]{wu2018beyond}
Lijun Wu, Xu~Tan, Di~He, Fei Tian, Tao Qin, Jianhuang Lai, and Tie-Yan Liu. 2018.
\newblock Beyond error propagation in neural machine translation: Characteristics of language also matter.
\newblock \emph{arXiv preprint arXiv:1809.00120}.

\bibitem[{Yang et~al.(2023)Yang, Li, and Liu}]{yang2023failures}
Zeyuan Yang, Peng Li, and Yang Liu. 2023.
\newblock Failures pave the way: Enhancing large language models through tuning-free rule accumulation.
\newblock In \emph{Proceedings of the 2023 Conference on Empirical Methods in Natural Language Processing}, pages 1751--1777.

\bibitem[{Yao et~al.(2022)Yao, Zhao, Yu, Du, Shafran, Narasimhan, and Cao}]{yao2022react}
Shunyu Yao, Jeffrey Zhao, Dian Yu, Nan Du, Izhak Shafran, Karthik Narasimhan, and Yuan Cao. 2022.
\newblock React: Synergizing reasoning and acting in language models.
\newblock \emph{arXiv preprint arXiv:2210.03629}.

\bibitem[{Yuan et~al.(2022)Yuan, Coenen, Reif, and Ippolito}]{yuan2022wordcraft}
Ann Yuan, Andy Coenen, Emily Reif, and Daphne Ippolito. 2022.
\newblock Wordcraft: story writing with large language models.
\newblock In \emph{27th International Conference on Intelligent User Interfaces}, pages 841--852.

\bibitem[{Zhang et~al.(2024)Zhang, Jiang, Liu, Chen, and Zhang}]{zhang2024llm}
Meishan Zhang, Gongyao Jiang, Shuang Liu, Jing Chen, and Min Zhang. 2024.
\newblock Llm--assisted data augmentation for chinese dialogue--level dependency parsing.
\newblock \emph{Computational Linguistics}, pages 1--24.

\bibitem[{Zhao et~al.(2024)Zhao, Huang, Xu, Lin, Liu, and Huang}]{zhao2024expel}
Andrew Zhao, Daniel Huang, Quentin Xu, Matthieu Lin, Yong-Jin Liu, and Gao Huang. 2024.
\newblock Expel: Llm agents are experiential learners.
\newblock In \emph{Proceedings of the AAAI Conference on Artificial Intelligence}, volume~38, pages 19632--19642.

\bibitem[{Zheng et~al.(2024{\natexlab{a}})Zheng, Chiang, Sheng, Zhuang, Wu, Zhuang, Lin, Li, Li, Xing et~al.}]{zheng2024judging}
Lianmin Zheng, Wei-Lin Chiang, Ying Sheng, Siyuan Zhuang, Zhanghao Wu, Yonghao Zhuang, Zi~Lin, Zhuohan Li, Dacheng Li, Eric Xing, et~al. 2024{\natexlab{a}}.
\newblock Judging llm-as-a-judge with mt-bench and chatbot arena.
\newblock \emph{Advances in Neural Information Processing Systems}, 36.

\bibitem[{Zheng et~al.(2024{\natexlab{b}})Zheng, Zhang, Zhang, Ye, Luo, and Ma}]{zheng2024llamafactory}
Yaowei Zheng, Richong Zhang, Junhao Zhang, Yanhan Ye, Zheyan Luo, and Yongqiang Ma. 2024{\natexlab{b}}.
\newblock \href {https://arxiv.org/abs/2403.13372} {Llamafactory: Unified efficient fine-tuning of 100+ language models}.
\newblock \emph{Preprint}, arXiv:2403.13372.

\bibitem[{Ziems et~al.(2024)Ziems, Held, Shaikh, Chen, Zhang, and Yang}]{ziems2024can}
Caleb Ziems, William Held, Omar Shaikh, Jiaao Chen, Zhehao Zhang, and Diyi Yang. 2024.
\newblock Can large language models transform computational social science?
\newblock \emph{Computational Linguistics}, 50(1):237--291.

\end{thebibliography}

\appendix

\section{Prompts for LLMs}
\label{sec:prompts}
We list all prompts in Table \ref{tab:prompts}.
All prompts follow a similar format.
First, we assign a role to each LLM agent by a sentence.
We then specify the task and background in one or two sentences.
Next, we give each LLM step-by-step instructions.
After that, we input the rules to be followed for each LLM.
Finally, the format of the output is specified as a ``markdown'' style to facilitate the extraction and support the information flow among LLMs.
In our preliminary study, this pattern works well in prompting various LLMs, with strong task completion performance and format adherence.

\section{Failed Attempts}
\label{sec:fail}
This appendix outlines our unsuccessful attempts. 
We hope that it will help follow-up research.
First, as a commonly used mechanism in LLM agents, reflection can support iterative enhancement by consolidating prior experiences \citep{yao2022react, park2023generative, yang2023failures}. 
In our pilot experiments, however, this approach did not succeed in refining the journalist's writing.
A potential explanation could be that the ASJ task for general audience requires specific revision instructions, whereas solely summarizing prior writing experiences results in general guidance only.

Within our framework, an LLM acts as a reader, reading an article and taking notes.
How about having this reader perform a reading comprehension task instead of simply taking notes?
Intuitively, it can demonstrate the reader's understanding and induce content complexity.
Our preliminary investigations, however, revealed that the reading comprehension approach yields less efficacy compared to the note-taking strategy.
It might be that the quality of question generation greatly affects the efficiency of the reading comprehension results.
Also, asking a fixed number of questions narrows the space of textual exploration, thereby restricting a comprehensive perception of content complexity.

We have tried to let an LLM (either Qwen 1.5 or GPT-3.5) do the writing-reading-editing task with a single prompt, but the LLM did not seem to follow the instructions successfully, e.g., it generated writing without giving or taking any feedback. 
We suspect that these models have limited ability to follow complex instructions.

Next, we briefly discuss the modules in our framework that may go wrong at runtime.
We observe that, in some cases, one of the LLMs did not follow editing instructions but just simply copied the input to the generated article.
This problem may be solved as the capabilities of the model increase.

\section{Datasets and Hyperparameters}
\label{sec:hyper}
Table \ref{tab:stat} shows some statistics of the three datasets used in our experiments.
\emph{SCITech(News)} is released by \citet{cardenas2023don}, who gathered press releases from ACM Technews as well as their source articles from various publishers, involving fields of computer science, engineering, astrophysics, biology, and others. 
\emph{eLife} is an open-access journal that focuses on biomedical and life sciences. 
\citet{goldsack2022making} collected some eLife articles as well as digests written by journal editors based on both the article itself and questions answered by the author. 
Similarly, \emph{PLOS} hosts journals across areas of science and medicine. 
Some of these articles, also collected by \citet{goldsack2022making}, come with the author's summary. 
For resource saving, the original paper's abstract serves as the scientific content input.
The same as previous studies, in \emph{SCITech}, we use 1,431 instances for training and validation, and the remaining 1,000 for testing, and each of \emph{eLife} and \emph{PLOS} datasets is separated into training, validation, and testing splits at a ratio of 90\%/5\%/5\%.
\begin{table}[t]
    \small
    \centering
    \setlength{\tabcolsep}{9pt}
    \begin{tabular}{lrrrrr}
        \toprule
       Statistics & \textbf{SCITech} & \textbf{eLife} & \textbf{PLOS} \\
        \midrule
        \# $\text{pairs}$ & 2431 & 4828 & 27525 \\
        \# $\text{words}^\text{ori}$ & 216.8 & 166.3 & 268.3\\
        \# $\text{sentences}^\text{ori}$ & 5.7 & 6.8 & 10.2 \\
        \# $\text{words}^\text{pln}$ & 176.1 & 347.6 & 175.6 \\
        \# $\text{sentences}^\text{pln}$ & 7.9 & 15.7 & 7.8\\
        \bottomrule
    \end{tabular}
    \caption{Statistics of benchmark datasets. Numbers of words and sentences are average values. The ``ori'' superscript denotes abstracts of original papers, and ``pln'' represents plain summaries written by humans.}
    \label{tab:stat}
\end{table}

Our local LLM service runs on a machine with eight GTX 4090 GPUs.
We utilize the Huggingface platform \citep{wolf2019huggingface} for downloading and loading checkpoints.
For rapid inference and memory efficiency, we utilize the vLLM library\footnote{\url{https://docs.vllm.ai}} to develop API services.
We deploy agents from the Qwen-1.5 series, for their good performance and diverse model scales \cite{bai2023qwen}.
In particular, Qwen-1.5-7B is employed for the steps of writing, providing suggestions, and revision, whereas Qwen-1.5-1.8B serves as the reader for taking notes.
To improve memory efficiency, we implement activation-aware weight quantization (AWQ, \citealt{lin2023awq}) for model quantization.
For one-shot learning settings, we randomly take one sample from the parallel datasets as a demonstration injected in the prompt for the journalist LLM.
For fine-tuning, we utilize LoRA \citep{hu2021lora} with Llama-Factory \citep{zheng2024llamafactory}, adopting the default setting with the number of epochs set to 10.
We use the default temperature setting and empirically set top\_p to 0.4, both frequency penalty and repetition penalty to 1, ensuring the stability of the LLMs' output while retaining diversity. 
We iterate five times and empirically select the output from the third iteration as the final result, for a balance of performance and cost-effectiveness.
The maximum number of tokens in the model output is 4,096.
Our preliminary tuning showed that such hyperparameters exhibited a relatively good and robust performance in various LLMs.
For the balance of fairness and resource expenditure, all approaches in our comparison study share the same set of hyperparameters setting.

\section{Methods Under Comparison}
\label{sec:comp}
We list detailed descriptions of all methods under comparison in Table \ref{tab:comp}.

\begin{table*}[t]
    \small
    \centering
    \begin{tabular}{lp{11cm}}
    \toprule
    \textbf{Approach}  & \textbf{Description} \\
    \midrule
    LLaMA-2-7B, LLaMA-3-8B     &  Two generations of LLaMA \citep{touvron2023llama, llama3} are tested for comparison. We also adopt one-shot learning (OS) and fine-tuning (FT) on LLaMA-3-8B, as well as test these variants in our collaborative framework.\\
    % Gemma-7B     &  \\
    Mistral-7B, Mixtral-8x7B     &  We test two scales of Mistral \citep{jiang2023mistral} to investigate the impact of model size.  \\
    Qwen-1.5-7B, Qwen-1.5-72B  &    We use three implementations of Qwen-1.5-7B \cite{bai2023qwen} for testing, i.e., the original version, with one-shot demonstration, and after fine-tuning, for a systematic study. The 72B version is for testing the impact of model size.  \\
    GPT-3.5-Turbo, GPT-4   &  We test the performance of two closed LLMs, GPT-3.5-Turbo-1106 \citep{openaichatgpt}, and GPT-4-1106-preview \citep{openaichatgpt}. \\
    BART      &  Previous ASJ studies \citep{goldsack2022making, cardenas2023don} took BART \citep{LewisBart} as baselines. We also include it in our study. \\
    CollabStory    &  \citet{venkatraman2024collabstory} proposed a method that involves three individual LLMs to generate the beginning, middle, and end of stories in sequence. We adopt this collaborative paradigm for ASJ to compare with our framework. \\
    ChatDev     &  \citet{qian2023communicative} introduced an LLM-powered software development framework, where LLM-driven agents communicate with each other through a prompt-based workflow, including design, coding, and testing. We adapt this framework to ASJ by changing phases to outline design, writing, and reviewing and editing. As each phase involves an instructor and an assistant, there are six agents in total, doubling the number of our framework. We take Qwen-1.5-7B as each agent for a fair comparison.  \\
    JRE-L    &  We test three versions of our framework. The first is 2×Qwen1.5-7B+Qwen1.5-1.8B. The second is with the journalist replaced by a one-shot demonstrated version as a warm start (OS). The third uses a fine-tuned journalist (FT) as a warm start. \\
    \bottomrule
    \end{tabular}
    \caption{Description of methods under comparison.}
    \label{tab:comp}
\end{table*}

\section{Details of Human Evaluation}
\label{sec:questionnaire}
\begin{figure*}[t]
    \centering
    \includegraphics[width=1\textwidth]{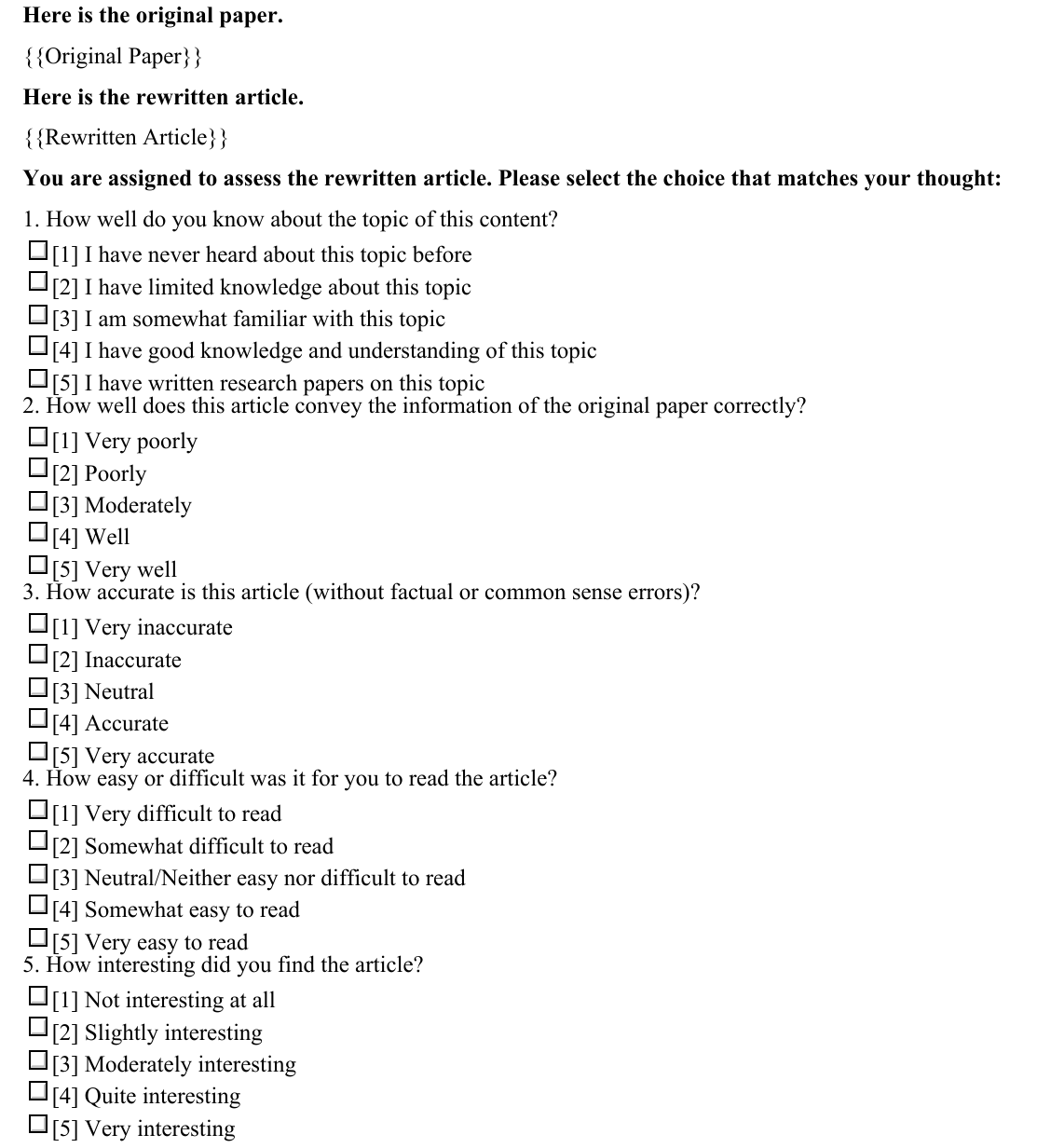}
    \caption{The questionnaire for participants to evaluate articles.}
    \label{fig:question}
\end{figure*}

Automatically assessing content authenticity and informativeness has been a challenging task.
\citet{cardenas2023don} used QuestEval \citep{Scialom_Dray_Lamprier_Piwowarski_Staiano_Wang_Gallinari_2021} to assess the faithfulness of ASJ-generated content, yet the results exhibited significant variances.
Therefore, human evaluation remains the main method for such assessments.
We created a questionnaire for human evaluation using a 1-5 Likert scale, as shown in Figure \ref{fig:question}. 
All participants were informed that their assessments would be used for research purposes.
We utilized Label Studio \citep{Label-Studio} to construct the annotation platform. 
Initially, participants indicated their familiarity with the given topic.
They were then tasked with answering four questions on Readability, Information Conveyance, Authenticity, and Interestingness respectively.
Readability assesses how easily the article can be read. 
Information conveyance determines how the rewritten content accurately and comprehensively conveys the information from the original paper. 
Similarly, authenticity assesses the correctness of the content, as a high-quality article should contain minimal factual or common sense errors. 
Finally, the level of interestingness is also a crucial factor; content of high appeal will provide a better reading experience.

% \section{Discussion on Metrics}
% \label{sec:disc}

% This is an appendix.

\section{Ablation Results}
\label{sec:abl_all}
Table \ref{tab:abl_all} presents the detailed results of the ablation study, demonstrating the effectiveness of our proposed framework. 
Our framework exhibits stability when component LLMs are replaced by different models. 
Additionally, upon the removal of one or more modules of our framework, the performance undergoes a significant decline, highlighting the efficacy of our framework design.
Moreover, we analyze the impact of changing LLMs to the latest Qwen version (2.5) and LLaMA (3.2) in a complementary experiment, as main studies including automatic evaluation and human evaluation have been finished before the latest models were released.
The result shows that newer models can improve the performance, but the improvement is limited, possibly due to marginality through the iterative process.
\begin{table*}[t]
\centering
\small
\setlength{\tabcolsep}{5.6pt}
\begin{tabular}{lcccccccccc}
\toprule
                                               & \multicolumn{3}{c}{\textbf{SCITech}}                                                                                 & \multicolumn{3}{c}{\textbf{eLife}}                                                                                  & \multicolumn{3}{c}{\textbf{PLOS}}                                                                                    &                                       \\ \cmidrule(lr){2-4} \cmidrule(lr){5-7} \cmidrule(lr){8-10}
\multirow{-2}{*}{\textbf{Approach}}            & \textbf{CLI}$\downarrow$                        & \textbf{FKGL}$\downarrow$                      & \textbf{DCRS}$\downarrow$                       & \textbf{CLI}$\downarrow$                    & \textbf{FKGL}$\downarrow$                   & \textbf{DCRS}$\downarrow$                    & \textbf{CLI}$\downarrow$                    & \textbf{FKGL}$\downarrow$                    & \textbf{DCRS}$\downarrow$                 & \multirow{-2}{*}{\textbf{Avg.}$\downarrow$}       \\ 
\midrule
JRE-L      & {\textbf{12.69}} & {\textbf{10.16}} & {\underline{9.79}} & {\textbf{11.60}} & 10.10                                & \underline{9.46}                                 & {12.74} & {\underline{10.00}} & {\textbf{9.69}} & {\textbf{10.69}}   \\
Reader$\rightarrow$7B  & 12.81 & 10.35 & \textbf{9.68} & \underline{11.82} & \underline{10.01} & 9.51 & \underline{12.67} & \textbf{9.93} & \underline{9.78} & 10.73 \\
LLM$\rightarrow$LLaMA  & \underline{12.77} & \underline{10.24} & 9.97  & 11.95 & \textbf{9.83} & \textbf{9.34} & \textbf{12.51} & 10.07 & 9.84  & \underline{10.72}   \\
$-$ Reading	& 13.21 & 10.63 & 10.33 & 12.22 & 10.78 & 10.02 & 13.35 & 10.59 & 10.25 & 11.26 \\
$-$ Suggestions	& 13.25 & 10.69 & 10.39 & 12.17 & 10.83 & 10.08 & 13.31 & 10.74 & 10.42 & 11.32 \\
$-$ Collaboration	& 13.50 & 11.01 & 10.71 & 12.47 & 10.99 & 10.41 & 13.65 & 10.91 & 10.70 & 11.59 \\
\midrule
LLM$\rightarrow$Qwen-2.5  &   12.51  &  10.02  &  \textbf{9.52}   &  11.42  &  \textbf{9.89}   &  9.23   &  12.53  &  9.81   &  \textbf{9.53}    &  10.50  \\
LLM$\rightarrow$LLaMA-3.2  &   \textbf{12.37} & \textbf{9.88} & 9.56 & \textbf{11.21} & 9.95 & \textbf{9.01} & \textbf{12.42} & \textbf{9.64} & 9.61 & \textbf{10.41}  \\
\bottomrule
    \end{tabular}
    \caption{Results of the ablation study.}
    \label{tab:abl_all}
\end{table*}

\section{Use of AI Assistants}
We use ChatGPT for correcting grammar and improving expressions in this manuscript.

\begin{table*}[t]
\centering
\small
\begin{tabular}{p{15.56cm}}
\toprule
\textbf{Journalist.}
You are a science journalist for general audiences. Given a paper's summary, you are assigned to rewrite it into a short understandable article for general audiences.\\
Follow the rules strictly: \\
- Keep short yet informative.\\
- The output format:\\
\#\# Article\\
...\\
\midrule
\textbf{Reader.}
You are a general reader. Given a popular science article, please read it carefully and take some notes.\\
Please take the following steps:\\
1. First, extract all technical terms with their context from the article.\\
2. Then, explain the technical terms based on their context.\\
Follow the rules strictly:\\
- Extraction should mention the specific location of each technical term in the article.\\
- Explanation should be first extracted from the article; if not found, it can be some common-sense or specialized knowledge.\\
- Extraction and explanation should be in points, like "1...2...3...".\\
- The output format:\\
\#\#\# Extraction\\
1. ...\\
2. ...\\
...\\
\#\#\# Explanation\\
1. ...\\
2. ...\\
\midrule
\textbf{Editor.}
You are a senior editor. Here are a scientific paper summary, and a short popular science article. A general reader has read the science article and takes some notes.\\
Please take the following steps:\\
1. First, evaluate the **reader's notes** based on these factors: content accuracy, lexical and technical complexity, and information conveyance (from the original content). \\
2. Then, based on the above evaluation, list some brief yet informative writing advice that may benefit the popular science article, to make the article easier for general readers without specialized knowledge to read and understand. Specifically, the advice should benefit these factors of the article:\\
    a) Content Accuracy: The factual correctness, scientific validity, and absence of errors in the general popular science article. \\
    b) Accessibility: Higher accessibility means less technical, more readable and interesting, etc. \\
    c) Information Conveyance: How effectively key information from the original paper is transferred to the popular science article.\\
Follow the rules strictly: \\
- Evaluation and advice sections should be in points, like "1...2...3...".\\
- Each advice should not go beyond the fact of original paper, but can be some common-sense or specialized knowledge.\\
- Each advice should be targeted at one specific aspect of the article.\\
- Don't suggest visualization, references or links.\\
- Suggest explanations rather than content additions.\\
- The output format: \\
\#\# Evaluation for reader's notes\\
- Content accuracy of reader's notes: ...\\
- Lexical and technical complexity of reader's notes: ...\\
- Information conveyance of reader's notes: ...\\
\#\# Advice\\
1. ...\\
2. ...\\
\midrule
\textbf{Revision}
You are a science journalist for general audiences. Given the paper summary and a short summary of the popular science article, you are assigned to rewrite the popular science summary for general audiences, who have no specialized knowledge. You are given some writing advice.\\
Please take the following steps: \\
1. Choose and refine the most relevant and suitable advice for writing improvement.\\
2. Then, based on the refined advice and the paper summary, rewrite the popular science article.\\ 
Follow the rules strictly: \\
- Keep your article short yet informative.\\
- Don't include visualization, references or links.\\
- Revision must not go beyond the original paper, but can be with some additional common-sense or professional knowledge for explanation.\\
- The output format:\\
\#\# Improvement\\
...\\
\#\# Revised Article\\
...\\
\bottomrule
\end{tabular}
\caption{Prompts for each LLM agent.} 
\label{tab:prompts}
\end{table*}

\end{document}